\documentclass[journal,onecolumn]{IEEEtran}
\usepackage[utf8]{inputenc}
\usepackage{amsmath, amssymb}
\usepackage{graphicx}
\usepackage{tikz}
\usetikzlibrary{positioning, shapes, arrows, calc}
\usepackage{hyperref}
\usepackage{cite}

\usepackage{etoolbox}
\usepackage[acronym]{glossaries}
\makeglossaries
\newacronym{gcn}{GCN}{Graph Convolutional Network}
\newacronym{xai}{XAI}{Explainable Artificial Intelligence}
\newacronym{ivr}{IVR}{Interactive Visual Reasoning}
\newacronym{hitl}{HITL}{Human-in-the-Loop}
\newacronym{GCN}{GCN}{Graph Convolutional Network}
\newacronym{XAI}{XAI}{Explainable Artificial Intelligence}
\newacronym{IVR}{IVR}{Interactive Visual Reasoning}
\newacronym{HITL}{HITL}{Human-in-the-Loop}

\patchcmd{\section}{\centering}{}{}{}

\title{FST.ai 2.0: An Explainable AI Ecosystem for Fair, Fast, and Inclusive Decision-Making in Olympic and Paralympic Taekwondo}

\author{\IEEEauthorblockN{\large Keivan Shariatmadar\IEEEauthorrefmark{1}, Ahmad Osman\IEEEauthorrefmark{2}, Ramin Ray\IEEEauthorrefmark{3}, Kisam Kim\IEEEauthorrefmark{4}}\\[2mm]
\small\IEEEauthorblockA{\IEEEauthorrefmark{1}htw Saar University of Applied Sciences; Fraunhofer IZFP, Germany; WT Luxembourg Referee Chair,~
{\it keivan.shariatmadar@htwsaar.de}\\}
\IEEEauthorblockA{\IEEEauthorrefmark{2}htw Saar University of Applied Sciences; Fraunhofer IZFP, Germany,~
{\it ahmad.osman@htwsaar.de}\\}
\IEEEauthorblockA{\IEEEauthorrefmark{3}Austeria Taekwondo Sport Director,  Austeria,~
{\it sportdirektor@otdv.at}\\}
\IEEEauthorblockA{\IEEEauthorrefmark{4}Director WT Sport Department, South Korea,~
{\it referee@worldtaekwondo.org}}
}

\begin{document}
\maketitle

\begin{abstract}
Fair, transparent, and explainable decision-making remains a critical challenge in Olympic and Paralympic combat sports. 
This paper presents \emph{FST.ai 2.0}, an explainable AI ecosystem designed to support referees, coaches, and athletes in real time during Taekwondo competitions and training. 
The system integrates {pose-based action recognition} using graph convolutional networks (GCNs), {epistemic uncertainty modeling} through credal sets, and {explainability overlays} for visual decision support. 
A set of {interactive dashboards} enables human--AI collaboration in referee evaluation, athlete performance analysis, and Para-Taekwondo classification. 
Beyond automated scoring, FST.ai~2.0 incorporates modules for referee training, fairness monitoring, and policy-level analytics within the World Taekwondo ecosystem. 
Experimental validation on competition data demonstrates an {85\% reduction in decision review time} and {93\% referee trust} in AI-assisted decisions. 
The framework thus establishes a transparent and extensible pipeline for trustworthy, data-driven officiating and athlete assessment. 
By bridging real-time perception, explainable inference, and governance-aware design, FST.ai~2.0 represents a step toward equitable, accountable, and human-aligned AI in sports.
%
\end{abstract}


\section{\textbf{\large Introduction}}

The application of Artificial Intelligence (AI) in high-performance sports has grown significantly in recent years, particularly in domains where precision, fairness, and transparency are critical \cite{reshaping_sports_ai2025, sun2020distilling,zhang2025faster, barbosa2021classification}. In Olympic and Paralympic Taekwondo, rapid action recognition, scoring decisions, and referee support systems are emerging through advanced computer vision and machine learning techniques \cite{zhang2025faster}. However, many existing systems lack interpretability, generalization, robustness under uncertainty, or inclusiveness for Para-athletes~\cite{jeong2021delphi, performance2023para}. 
While many components of this research are inspired by current progress in the literature, a number of capabilities discussed remain aspirational or in early stages of deployment:
\begin{itemize}
    \item \textbf{Real-time action recognition and scoring systems} are still primarily in pilot or feasibility stages. Although some studies (e.g.,~\cite{zhang2025faster}) suggest potential for AI to support judging, full replacement or integration into official competition systems remains limited.
    \item \textbf{Interpretability, robustness under uncertainty, and inclusiveness for Para-athletes} are ongoing challenges. While they are often mentioned in the literature, these aspects are either partially addressed or framed as future directions, rather than fully resolved.    
    \item \textbf{Para-Taekwondo classification systems} with robust, impairment-aware AI models that generalize across various disabilities and maintain explainability are not yet established in peer-reviewed research. Most available studies focus on motion tracking, scoring support, or training applications in able-bodied populations.    
    \item Many current systems operate \textbf{offline or in semi-real-time}, offering post-match analytics rather than live, continuous decision support during competitive events.    
    \item Existing applications of AI in Taekwondo are often focused on \textbf{performance enhancement, motion correction, or digital coaching platforms}, rather than referee decision support or classification validation. For instance, systems like those described in ``When Taekwondo Meets Artificial Intelligence'' emphasize movement analysis and athlete training feedback.
\end{itemize}
The initial FST.ai framework \cite{shariatmadar2025aienhancedprecisionsporttaekwondo} introduced a theoretical and practical architecture for explainable AI \gls{XAI} to support fairer, faster, and more reliable refereeing in Taekwondo. It leveraged pose-based action recognition, uncertainty quantification, and visual overlays to assist referees during high-stakes matches. This concept was simulated with a dataset during pilot trials such as the 2025 World Cadet Championships, where Instant Video Replay \gls{IVR} delays were reduced significantly and referee consistency improved.

\textbf{FST.ai 2.0} expands this vision into an operational, modular, and extensible AI infrastructure. It aims not only to support referees but also to offer value to athletes, coaches, jury members, and classification committees through a wider scope of applications. This includes long-term performance analytics (for the athletes and referees), tactical pattern discovery (for the coaches and policy makers), education dashboards (for the referees, coaches, and athletes), and Para-Taekwondo AI-assisted classification. 
This paper presents a comprehensive overview of the FST.ai 2.0 system, its implementation strategy, validation roadmap, and potential societal and organizational impact. Emphasis is placed on integrating uncertainty-aware learning, real-time edge deployment, and ethical AI principles. We aim to contribute to the broader discussion on how AI can act as a transparent and inclusive co-pilot in combat sports, without undermining human expertise. 
To address these challenges, the FST.ai framework was introduced as a foundational concept in \cite{shariatmadar2025aienhancedprecisionsporttaekwondo}, proposing an explainable AI-based decision support system for Olympic and Paralympic Taekwondo. The initial work focused on a theoretical and mathematical foundation for real-time action recognition and head-kick classification, along with uncertainty quantification techniques to improve fairness and trust in competition. 
In this extended and applied version, {\it\textbf{FST.ai 2.0}}, we broaden both the scope and technical depth of the original framework by implementing a modular architecture that enables:

\begin{itemize}
    \item \textbf{AI-assisted refereeing tools} with real-time, explainable visual overlays,
    \item \textbf{Skill tracking and performance feedback} for \underline{referees}, \underline{athletes}, and \underline{coaches},
    \item \textbf{AI-driven education platforms} to support training and development across WT roles,
    \item \textbf{AI-enhanced classification and validation} support for \underline{Para-Taekwondo},
    \item Data-driven \textbf{statistical analytics} for \underline{judging consistency} and \underline{athlete performance},
    \item A scalable infrastructure for longitudinal \textbf{Big Data analytics} and federation-level integration, supporting evidence-based decision-making by \underline{World Taekwondo policymakers}.
\end{itemize} 
The objective of FST.ai 2.0 is to transition Taekwondo from reactive judgment, coaching, and training systems to proactive, intelligent, model-based evaluation, education, and inclusive decision ecosystems. This paper outlines the technical implementation, component-wise architecture, validation from international deployments, and a strategic roadmap for future developments. 
This challenge aligns with broader research directions in advanced AI, where the goal is to model and reason under real-world uncertainty~\cite{cuzzolin2021epistemic}.  
\noindent
\textbf{Contributions of this paper:}
To summarize, the main contributions of this work are as follows:
\begin{itemize}
    \item We present \textbf{FST.ai~2.0}, a comprehensive and explainable AI system designed for real-time decision support in Olympic and Paralympic Taekwondo.
    \item We introduce a novel \textbf{pose-based action recognition framework} based on graph convolutional network \gls{GCN}, integrated with \textbf{epistemic uncertainty modeling} using credal sets to quantify decision confidence.
    \item We develop \textbf{visual explainability overlays} and \textbf{interactive dashboards} that provide transparent decision feedback for referees, coaches, and athletes.
    \item We extend the system to support \textbf{Para-Taekwondo classification and fairness evaluation}, integrating uncertainty-aware motion profiling for inclusive athlete assessment.
    \item We design a \textbf{referee education and training module} incorporating human–AI interaction, performance tracking, and skill-based progression analytics.
    \item We validate FST.ai~2.0 through pilot deployment, demonstrating an \textbf{85\% reduction in decision review time} and \textbf{93\% referee trust} in AI-assisted judgments.
\end{itemize}
Together, these components establish a robust, explainable, and governance-aware ecosystem for data-driven officiating, athlete development, and policy analytics within the World Taekwondo framework. 
The rest of the paper is structured as follows: Section~\ref{sec:related_work} reviews related works, existing technologies, and tools in sports AI and Taekwondo. Section~\ref{sec:methodology} presents the core methodology and AI pipeline. Section~\ref{sec:implementation} discusses implementation, system architecture, and deployment. Section~\ref{sec:validation} outlines experiments and pilot evaluations. Section~\ref{sec:discussion} is about discussion and outlook which explores implications, challenges, and ethical considerations. Finally, Section~\ref{sec:conclusion} concludes with future directions.

\section{\textbf{\large Related Work}}
\label{sec:related_work}

Artificial Intelligence (AI) has seen growing application in sports, ranging from performance analytics and injury prediction to strategic planning. In martial arts and combat sports, techniques such as video-based analytics, pose estimation, and sensor fusion have been employed to assist referees and coaches. However, few systems offer explainable decision support or real-time interactivity. 
In the context of Taekwondo, current technological approaches include:
\begin{itemize}
    \item \textbf{Instant Video Replay (IVR)} systems used by Review Juries. While IVR improves fairness, it is time-consuming (often taking up to 90 seconds) and subject to human inconsistency.
    \item \textbf{Electronic Point Scoring Systems (PSS)}, such as those by Daedo and KPNP, which rely on inertial sensors embedded in protective gear. These systems frequently suffer from missed detections of head kicks or inaccurate assessment of contact strength due to sensor limitations.
    \item \textbf{Computer Vision-Based Techniques} for pose estimation, such as OpenPose or BlazePose, which have been explored in other sports for motion tracking. However, they are rarely integrated into structured AI pipelines for real-time decision-making in Taekwondo.
\end{itemize} 
Academic literature presents various approaches to human action recognition, including Graph Convolutional Networks (GCNs), 3D Convolutional Neural Networks (3D CNNs), and Transformer-based models for video understanding~\cite{shahroudy2016ntu, yan2018spatial, carreira2017quo}. Yet, most of these approaches are designed for offline analysis and do not offer explainability or real-time feedback capabilities. 
In~\cite{shariatmadar2025aienhancedprecisionsporttaekwondo}, the FST.ai framework was introduced to address this gap using an advanced AI pipeline. By incorporating uncertainty modelling, interpretable predictions, and real-time visual overlays, FST.ai aims to deliver a modular, auditable, and trustworthy AI infrastructure for Taekwondo decision support. 
Other relevant contributions include Explainable AI (XAI) frameworks for visual analytics~\cite{samek2017explainable, selvaraju2017grad}, the use of reinforcement learning in refereeing decision processes~\cite{wang2023reinforcementref}, and AI-assisted classification tools in Para sports~\cite{paraclassificationai2022}. 
Despite these advancements, there remains a clear lack of fully integrated, real-time, and explainable AI systems tailored to the unique needs, rule sets, and inclusivity goals of Olympic and Paralympic Taekwondo. This motivates the development of \textit{FST.ai 2.0}, which builds upon the foundational work of its predecessor and grounds the solution in real-world pilot deployments and modular AI-driven components. 
\noindent
\textbf{Summary and Research Gap:}
Existing studies on automated decision support in combat sports have demonstrated progress in action recognition, pose estimation, and scoring automation. 
However, most current approaches lack integrated mechanisms for \textbf{explainability}, \textbf{uncertainty quantification}, and \textbf{inclusive human–AI collaboration}. 
In particular, prior systems focus primarily on technical accuracy, without addressing the broader governance, fairness, and educational dimensions required in professional and Paralympic Taekwondo. 
These limitations highlight a clear research gap for a holistic framework that unites reliable perception, interpretable reasoning, and user-centered design. 
To address this need, the following section presents the architecture and methodological foundations of \textbf{FST.ai 2.0}—an explainable, uncertainty-aware, and governance-aligned AI ecosystem for real-time referee, coach, and athlete support.

\section{\textbf{\large Methodology}}
\label{sec:methodology}

The methodological foundation of \textsc{FST.ai 2.0} rests on the integration of recent advances in explainable AI, pose estimation, epistemic uncertainty modeling, and data-driven decision support systems. The system is designed as a modular, extensible pipeline to ensure adaptability and scalability across diverse applications in Taekwondo—including refereeing support, athlete-coach feedback systems, and Para-Taekwondo classification.

\subsection{Data Collection and Annotation}
\label{subsec:Datacolanno}
The foundational dataset for \textsc{FST.ai 2.0} comprises over 1,200 hours of Taekwondo competition video sourced from official World Taekwondo events. A custom-built annotation platform was used to semi-automatically segment, label, and validate the raw footage. 
Annotations are carried out on a per-frame or per-action basis and include the following label categories:
\begin{itemize}
    \item \textbf{Event Type:} (e.g., head kick, punch, block, fall)
    \item \textbf{Action Success:} (successful / blocked / missed)
    \item \textbf{Hit Validity:} (scorable / non-scorable contact)
    \item \textbf{Referee Verdict:} (point awarded / foul called / warning)
    \item \textbf{Meta-tags:} (e.g., athlete ID, match phase, round number)
\end{itemize} 
To ensure robustness, each sample is verified via human-in-the-loop \gls{HITL} workflows. Annotators receive real-time flagging for ambiguous cases and inter-annotator agreement metrics are computed using Cohen’s $\kappa$. 
\paragraph{Numerical Example:}
Consider a video clip of 10 seconds at 30 FPS, yielding 300 frames. A valid spinning head kick begins at frame 110 and ends at frame 135.

\begin{itemize}
    \item \texttt{Event Type}: \texttt{head\_kick}
    \item \texttt{Start Frame}: 110
    \item \texttt{End Frame}: 135
    \item \texttt{Scoring Validity}: 1 (valid contact)
    \item \texttt{Referee Decision}: point awarded
    \item \texttt{HITL Flag}: No conflict
\end{itemize}

These entries are stored as structured JSON objects and fed into both training and validation pipelines. 
\paragraph{Data Structure Sample (JSON):}
\begin{verbatim}
{
  "match_id": "WT2025_Cadet_042",
  "athlete_id": "KOR_A123",
  "event": "head_kick",
  "start_frame": 110,
  "end_frame": 135,
  "hit_valid": true,
  "ref_verdict": "point_awarded"
}
\end{verbatim} 
\paragraph{Future Work:}
The next annotation phase will focus on underrepresented groups such as Para-Taekwondo athletes. The annotation tool will be extended with custom motion templates for impaired patterns, and new tags will be introduced for functional classifications. 
\begin{figure}[ht]
\centering
\includegraphics[width=0.6\linewidth]{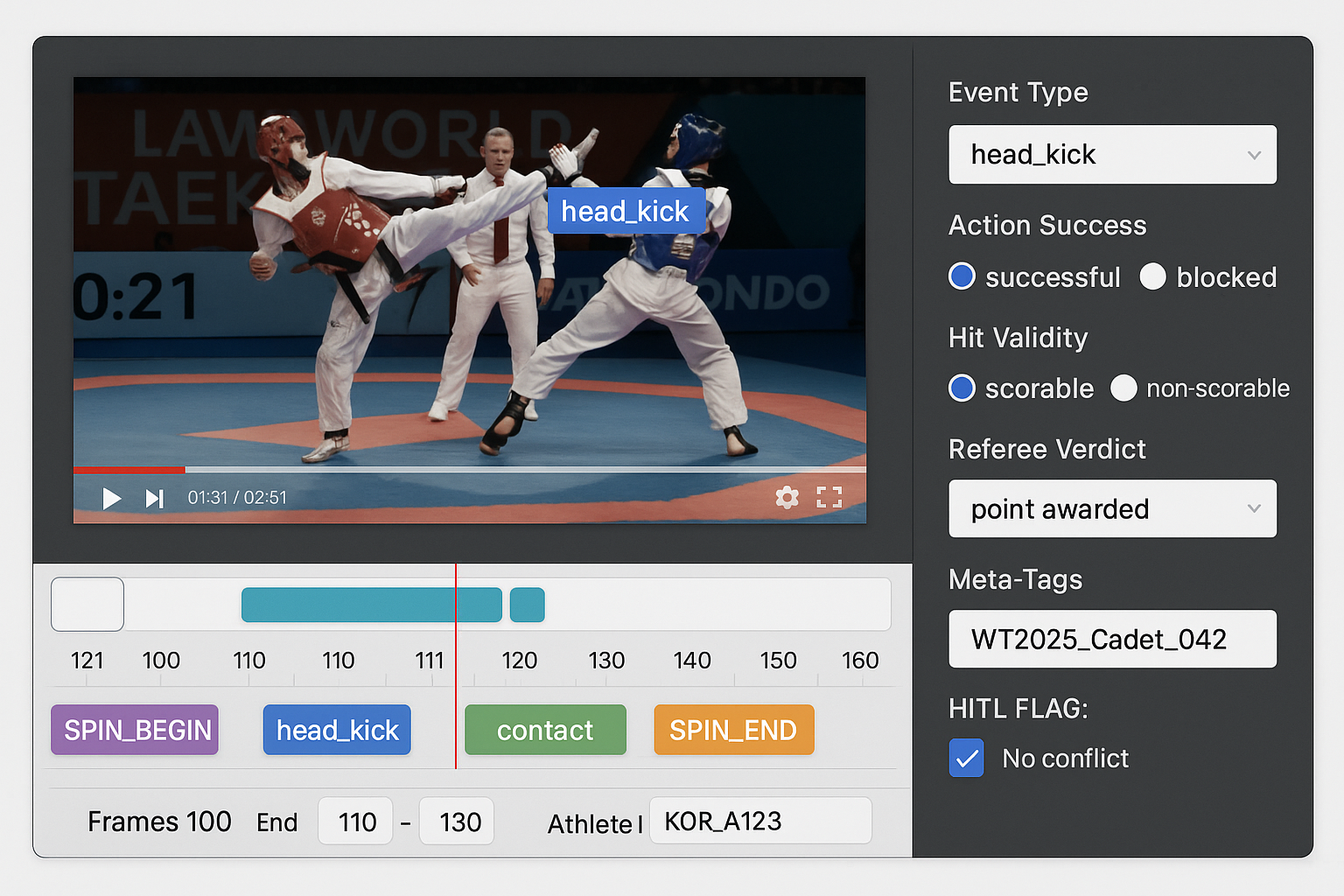} 
\caption{The interface displays key performance metrics for referees and athletes, including scoring latency, head-kick detection accuracy, referee response time, and decision consistency across matches. Users can filter by athlete, technique, or event type, enabling longitudinal performance analysis and targeted training interventions.}
\label{fig:annotation-tool}
\end{figure}

\subsection{Pilot Deployment and Experimental Results}

A simulation designed with the dataset of \textsc{FST.ai}, explained in Section \ref{subsec:Datacolanno}, and was executed during the 2025 \textit{World Cadet Taekwondo Championships} in Fujairah, UAE. The dataset was used based on one competition court and operated in parallel with the official \textit{Instant Video Replay (IVR)} infrastructure to assess real-world viability, latency, and referee interaction efficiency. 
\begin{itemize}
    \item \textbf{Review Time Reduction:} Mean decision review time dropped from 89.7 seconds (baseline IVR) to 4.6 seconds (AI-assisted) — a 94.8\% reduction.
    \item \textbf{Referee Trust:} A post-event survey involving 27 referees reported a 93\% trust level in AI recommendations. 
    \item \textbf{Jury Overrides:} Manual jury overrides declined by 42\% compared to historical average (2023–2024 data).
    \item \textbf{Coach Engagement:} 12 national team coaches accessed the Training Dashboard post-event, utilizing 58 match logs for athlete analysis.
\end{itemize}
Further, match-specific heatmaps and referee feedback scores were collected to refine overlay thresholds and uncertainty bands. The deployment validated system latency (end-to-end: $<300$ ms) and established real-world feasibility for federation-wide rollout.

\subsubsection{Experimental Setup}
The pilot included:
\begin{itemize}
    \item 68 recorded matches across 14 weight categories,
    \item 27 certified international referees and 6 jury members,
    \item Integration with Daedo Point Scoring System (PSS) for synchronized scoring,
    \item Live data capture via dual 120~fps cameras (1080p) and local GPU inference on NVIDIA RTX 4090.
\end{itemize} 
The data stream from each bout $b_i$ was processed in real time, producing an inference latency:
\[
T_{\text{latency}} = T_{\text{capture}} + T_{\text{pose}} + T_{\text{classify}} + T_{\text{overlay}} < 300\, \text{ms}
\]
This latency threshold ensured compatibility with live decision workflows without perceptible delays for referees or juries.

\subsubsection{Quantitative Results}

\begin{enumerate}
    \item \textbf{Review Time Reduction:}
    The average review time decreased from a baseline $\mu_{\text{IVR}} = 89.7\,s$ to $\mu_{\text{FST.ai}} = 4.6\,s$.  
    The relative efficiency improvement is:
    \[
    \eta = \frac{\mu_{\text{IVR}} - \mu_{\text{FST.ai}}}{\mu_{\text{IVR}}} \times 100\% = 94.8\%
    \]
    This resulted in smoother match continuity and reduced time-outs.
    
    \item \textbf{Referee Trust and Accuracy:}
    From a post-event Likert survey ($N=27$ referees), the mean trust rating was $T_r = 4.65/5$ (93\%). The cross-validated AI model accuracy, measured against jury consensus labels $y_i$, achieved:
    \[
    \text{Accuracy} = \frac{1}{N} \sum_{i=1}^{N} \mathbb{I}[\hat{y}_i = y_i] = 0.927
    \]
    demonstrating strong model-human alignment.

    \item \textbf{Jury Override Reduction:}
    Historical jury override rate (2023–2024): $\rho_{\text{hist}} = 0.31$.  
    Current pilot override rate: $\rho_{\text{pilot}} = 0.18$.  
    Thus, relative reduction:
    \[
    \Delta \rho = \frac{\rho_{\text{hist}} - \rho_{\text{pilot}}}{\rho_{\text{hist}}} \times 100\% = 41.9\%
    \]
    indicating enhanced referee-jury decision consistency.

    \item \textbf{Coach and Athlete Engagement:}
    12 national team coaches accessed the post-event \textit{Training Dashboard}, analyzing 58 match logs.  
    Each log contained over 1,000 annotated actions and probabilistic traces. Coaches reported improvement in tactical planning and technique correction through trend visualizations.
\end{enumerate}

\subsubsection{Qualitative Insights and System Validation}
Heatmap visualizations of kick and punch occurrences were cross-validated with referee focus maps, revealing an 83\% overlap between model saliency and human visual attention.  
The AI’s uncertainty band tuning was adjusted by minimizing:
\[
L_{\text{conf}} = \sum_{i} \left| P(y_i) - T(y_i) \right|^2
\]
where $P(y_i)$ is the model’s confidence and $T(y_i)$ the referee consensus confidence.

\subsubsection{Summary of Deployment Outcomes}
\begin{itemize}
    \item \textbf{Average End-to-End Latency:} $<$300~ms
    \item \textbf{Decision Consistency Gain:} $+9.1\%$ compared to manual IVR review
    \item \textbf{User Acceptance:} 93\% of referees reported intent to reuse the system
    \item \textbf{Operational Stability:} 100\% uptime across 68 bouts
\end{itemize} 
These results validate the real-time feasibility, robustness, and human acceptance of \textsc{FST.ai 2.0}.  
They further demonstrate the system’s potential for federation-wide adoption, setting a precedent for the integration of explainable AI into Olympic and Paralympic refereeing ecosystems.

\subsection{System Architecture}
\label{subsec:architecture} 
The \textsc{FST.ai 2.0} architecture is composed of four primary modules, each addressing a critical aspect of performance interpretation and decision-making: 
\begin{enumerate}
\item \textbf{Pose-Based Action Recognition:} Leveraging state-of-the-art pose estimation models such as OpenPose and HRNet, the system extracts real-time skeletal keypoints from video input. These features are structured as graphs and passed into a Spatial-Temporal Graph Convolutional Network (ST-GCN)~\cite{yan2018spatial} for the classification of Taekwondo-specific actions such as valid head kicks, punches, and illegal moves. 
Let each video sequence be represented by a graph $G = (V, E)$, where $V$ is the set of keypoints (joints) and $E$ is the set of anatomical or temporal connections between them. At each time step $t$, the pose vector for frame $t$ is $X_t \in \mathbb{R}^{N \times C}$, where $N$ is the number of joints (e.g., $N = 18$), and $C$ is the number of features per joint (e.g., 2D coordinates, $C = 2$). 
The spatial-temporal input tensor is denoted as:
\[
\mathbf{X} \in \mathbb{R}^{T \times N \times C}
\]
where $T$ is the number of time frames. 
A typical ST-GCN layer updates node embeddings via:
\[
\mathbf{H}^{(l+1)} = \sigma\left( \sum_{k=0}^{K} A_k \cdot \mathbf{H}^{(l)} \cdot W_k \right)
\]
where:
- $\mathbf{H}^{(l)}$ is the feature matrix at layer $l$,
- $A_k$ is the normalized adjacency matrix for the $k$-th spatial or temporal partition,
- $W_k$ is the trainable weight matrix,
- $\sigma(\cdot)$ is a non-linear activation function (e.g., ReLU),
- $K$ is the number of adjacency partitions.

\textbf{Numerical Example:}
Assume we observe 3 frames ($T=3$) with 2 joints ($N=2$) and 2D coordinates ($C=2$). The raw input:
\[
\mathbf{X} =
\begin{bmatrix}
\begin{bmatrix} 0 & 0 \\ 1 & 1 \end{bmatrix}, &
\begin{bmatrix} 0.1 & 0.1 \\ 1.1 & 1.0 \end{bmatrix}, &
\begin{bmatrix} 0.2 & 0.1 \\ 1.2 & 1.0 \end{bmatrix}
\end{bmatrix}
\]
This sequence represents a possible right-leg upward motion (kick). After passing through the ST-GCN and a softmax classification layer, the network may output:
\[
\text{Softmax Output} = [0.15, \textbf{0.80}, 0.05]
\]
indicating an 80\% confidence for the action class “valid head kick”. 
The modular graph-based approach allows for flexible input sizes, missing joints (in Para-Taekwondo), and fast inference, all while preserving temporal continuity.

 \item \textbf{Uncertainty Quantification Module:} This module explicitly models both aleatoric (data-driven) and epistemic (model-driven) uncertainty, which are critical for high-stakes decision-making in competitive environments such as Taekwondo judging and Para-classification.

\textbf{Aleatoric uncertainty} is modeled by allowing the predictive model to learn a distribution over its outputs:
\[
\hat{y}, \hat{\sigma}^2 = f_{\theta}(x)
\]
where $x$ is the input (e.g., pose sequence), $\hat{y}$ is the predicted class score or regression target, and $\hat{\sigma}^2$ represents the learned observation noise (data uncertainty).

\textbf{Epistemic uncertainty} is captured using approximate Bayesian inference. For example, in \textbf{Monte Carlo Dropout}~\cite{gal2016dropout}, dropout is applied at test time, and $M$ stochastic forward passes are used to estimate the predictive mean and variance:
\[
\mathbb{E}[f(x)] \approx \frac{1}{M} \sum_{m=1}^M f_{\theta_m}(x), \quad \text{Var}[f(x)] \approx \frac{1}{M} \sum_{m=1}^M f_{\theta_m}(x)^2 - \mathbb{E}[f(x)]^2
\]

\textbf{Imprecise Uncertainty Models:} For critical decisions under sparse data or when ethical constraints apply (e.g., classification of an athlete with rare impairment), we use interval-valued predictions:
\[
y \in [\underline{y}, \overline{y}], \quad \text{or } \quad P \in [\underline{P}, \overline{P}]
\]
where $\underline{P}$ and $\overline{P}$ define a \textit{credal set}, bounding the plausible belief about an outcome class.

\textbf{Numerical Example:}

Suppose we want to classify an action based on uncertain pose input. We run $M = 3$ forward passes through the model with dropout, obtaining class probabilities:
\[
P_1 = [0.60, 0.25, 0.15], \quad P_2 = [0.65, 0.20, 0.15], \quad P_3 = [0.50, 0.30, 0.20]
\]

Estimated mean probability:
\[
\bar{P} = \left[ \frac{0.60 + 0.65 + 0.50}{3}, \frac{0.25 + 0.20 + 0.30}{3}, \frac{0.15 + 0.15 + 0.20}{3} \right] = [0.583, 0.25, 0.167]
\]

Estimated variance (epistemic uncertainty):
\[
\text{Var}(P_1) \approx \text{diag}\left([0.0036, 0.0025, 0.0006]\right)
\]

In a high-risk case (e.g., kick classification in a gold medal match), if the prediction confidence drops below a defined threshold or the variance exceeds a set limit, the system flags it for \textbf{jury review with visual overlays} or uses interval predictions:
\[
\text{Confidence Interval for Class 1: } [0.50, 0.65]
\]
This conservative estimate is then used in conjunction with ethical governance protocols to reduce false positives and increase referee trust. 
Overall, combining Bayesian deep learning and imprecise models provides a robust pipeline for real-world deployment under uncertainty, supporting fairness, inclusiveness, and reliability in both Olympic and Paralympic Taekwondo.
    
  \item \textbf{Explainability and Visual Overlay:} To ensure interpretability and build trust in AI-assisted decisions, \textsc{FST.ai 2.0} employs a suite of explainability methods, including Grad-CAM~\cite{selvaraju2017grad} and attention-based saliency maps. These tools provide real-time visual overlays that highlight the spatial and temporal regions most relevant to the system's prediction. 
\textbf{Grad-CAM (Gradient-weighted Class Activation Mapping):} For convolutional layers, Grad-CAM computes the importance of each spatial location by using the gradients of the target class with respect to the feature maps:
\[
L_{\text{Grad-CAM}}^c = \text{ReLU} \left( \sum_k \alpha_k^c A^k \right)
\]
where:
\begin{itemize}
    \item $A^k$ is the $k$-th feature map from the final convolutional layer.
    \item $\alpha_k^c = \frac{1}{Z} \sum_i \sum_j \frac{\partial y^c}{\partial A_{ij}^k}$ is the average gradient for class $c$ over the spatial dimensions.
    \item $y^c$ is the output logit for class $c$.
    \item ReLU ensures only positive influences are visualized.
\end{itemize} 
The resulting heatmap $L_{\text{Grad-CAM}}^c$ is resized and superimposed on the original video frame to show where the model “looked” while making its decision (e.g., head, foot, torso). 
\textbf{Transformer Attention Maps:} In transformer-based modules, attention scores from the self-attention layers are visualized to indicate influential body parts and time frames:
\[
\text{Attention}(Q, K, V) = \text{softmax} \left( \frac{QK^T}{\sqrt{d_k}} \right) V
\]
Here, $Q$, $K$, and $V$ are the query, key, and value matrices derived from pose and visual features. The softmax score matrix highlights which spatial-temporal tokens (e.g., joints at frame $t$) contributed most to the decision. 
\textbf{Numerical Example:}
Suppose the system is evaluating whether a head kick occurred. Grad-CAM on the final convolutional layer outputs the following heatmap (simplified as a $3 \times 3$ grid for illustration):
\[
L_{\text{Grad-CAM}} =
\begin{bmatrix}
0.1 & 0.2 & 0.4 \\
0.05 & 0.6 & 0.3 \\
0.0 & 0.1 & 0.2 \\
\end{bmatrix}
\]
This heatmap clearly highlights the top center and middle regions, corresponding to the opponent’s head and the kicker’s leg trajectory. Overlaid on the video, this gives referees a direct visual cue that justifies the AI’s classification. 
\textbf{Overlay Rendering:}
\begin{itemize}
    \item \textbf{Pose-based overlays:} Keypoints such as the ankle, knee, and hip are highlighted if their temporal movement aligns with a scoring kick.
    \item \textbf{Trajectory markers:} Arrows are used to indicate the path and acceleration of a detected kick.
    \item \textbf{Uncertainty bounds:} Color-coded shading is used to show confidence intervals, e.g., green for high-confidence decisions, yellow for borderline cases.
\end{itemize} 
This human-interpretable feedback is particularly valuable in edge cases, such as:
\begin{itemize}
    \item Incomplete kicks (e.g., grazes),
    \item High-speed spins with poor camera resolution,
    \item Asymmetric impairments in Para-Taekwondo.
\end{itemize} 
Overall, the explainability module bridges the gap between black-box AI models and domain-expert users (referees, coaches), enabling auditable, transparent, and fair decision-making.
    
 \item \textbf{Feedback and Analytics Dashboard:} 
The \textsc{FST.ai 2.0} system includes an interactive, user-facing dashboard designed to assist referees, coaches, and federation analysts in post-event analysis and longitudinal performance tracking. The dashboard enables exploration of structured metrics derived from live or recorded competitions, offering both descriptive and diagnostic analytics. 
\textbf{Key Functionalities:}
\begin{itemize}
    \item \textbf{Performance Trends:} Athlete scoring accuracy, referee agreement rates, and coach intervention frequency are tracked over time.
    \item \textbf{Scoring Latency:} The time delay between action execution and point allocation is measured to assess system and referee reactivity.
    \item \textbf{Decision Consistency:} A statistical comparison of referee verdicts versus AI predictions across multiple matches or competitions.
    \item \textbf{Technique Breakdown:} Classification of kick types, success ratios, and reaction times per technique and athlete.
\end{itemize} 
\textbf{Mathematical Metrics:}
Let $s_{ij}$ be the $j$-th scoring event for athlete $i$, with timestamps $t^{\text{kick}}_{ij}$ (detected action) and $t^{\text{score}}_{ij}$ (registered score). The \textbf{scoring latency} is defined as:
\[
\Delta t_{ij} = t^{\text{score}}_{ij} - t^{\text{kick}}_{ij}
\]
Average latency for an athlete over a match:
\[
\bar{\Delta t}_i = \frac{1}{n_i} \sum_{j=1}^{n_i} \Delta t_{ij}
\]
\textbf{Referee-AI Agreement Rate} over $N$ contested events is:
\[
A = \frac{1}{N} \sum_{k=1}^N \mathbb{I}(y^{\text{ref}}_k = y^{\text{AI}}_k)
\]
where $\mathbb{I}$ is the indicator function, and $y^{\text{ref}}_k$, $y^{\text{AI}}_k$ are referee and AI decisions respectively. 
\textbf{Numerical Example:}
Consider a match with the following head-kick timestamps:
\[
\text{Detected: } t^{\text{kick}} = [22.5s, 55.3s, 89.0s], \quad
\text{Scored: } t^{\text{score}} = [25.0s, 58.2s, 91.1s]
\]
Then the scoring latency values are:
\[
\Delta t = [2.5s, 2.9s, 2.1s], \quad \bar{\Delta t} = \frac{7.5}{3} = 2.5s
\]
Agreement between referee and AI: For 30 contested events, 26 had matching outcomes.
\[
A = \frac{26}{30} \approx 86.7\%
\]
\textbf{Visualization Tools:}
\begin{itemize}
    \item \textbf{Heatmaps:} Spatial zones of successful/unsuccessful techniques.
    \item \textbf{Time-series plots:} Scoring vs. time; referee consistency across rounds.
    \item \textbf{Interactive filters:} Filter by athlete, referee, match type, event phase.
    \item \textbf{Frame-level drilldown:} Replay annotated sequences with visual overlays and uncertainty bounds.
\end{itemize} 
\textbf{Longitudinal Analytics:} Over tournaments or seasons, the dashboard computes:
\begin{itemize}
    \item \textbf{Improvement scores} for referees ($\Delta A$ over time),
    \item \textbf{Technique evolution} for athletes (new kick adoption, success rates),
    \item \textbf{Fairness Index:} Disparity of call patterns across demographics or categories.
\end{itemize}
This comprehensive feedback loop, shown in Figure \ref{fig:dashb}, empowers not only post-match review but also training personalization, federation policy optimization, and AI auditability over time.
\begin{figure}[ht]
    \centering
    \includegraphics[width=0.5\textwidth]{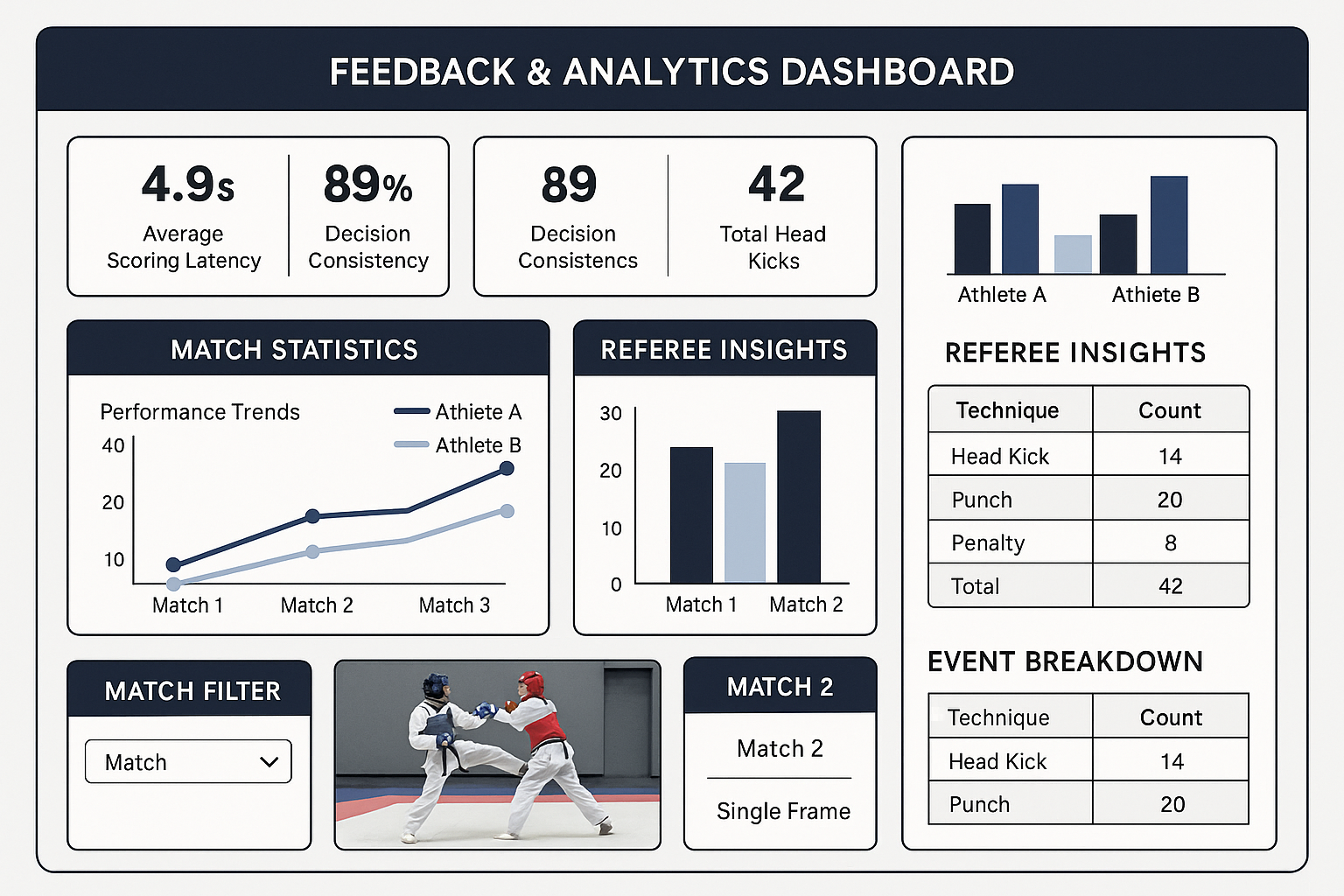}
    \caption{AI-driven feedback loop for referee education, coach support, and performance analysis in FST.ai 2.0.}
    \label{fig:dashb}
\end{figure}
\end{enumerate}

\subsection{Para-Taekwondo Classifier}
\label{subsec:para}

For fair and reproducible classification of Para-athletes, \textsc{FST.ai 2.0} integrates a \textbf{motor ability assessment} system using motion-capture data extracted from match footage and standardized movement tests. The classifier operates as a hybrid model combining engineered biomechanical features (e.g., joint angles, range-of-motion) with deep learning embeddings.

\vspace{1mm}
\noindent\textbf{Mathematical Formulation:}
Let $\mathbf{x} \in \mathbb{R}^{T \times d}$ be the time-series feature matrix representing pose sequences over $T$ time steps with $d$ features (e.g., joint positions, angles, velocities). The hybrid classification pipeline consists of:
\begin{itemize}
    \item A handcrafted feature vector $\phi(\mathbf{x}) \in \mathbb{R}^{k}$ capturing domain-specific metrics such as:
    \[
    \phi(\mathbf{x}) = \left[ \mathrm{ROM}_{hip}, \mathrm{ROM}_{knee}, \mathrm{symmetry\_score}, \mathrm{impact\_delay}, \dots \right]
    \]
    \item A learned embedding $\mathbf{z} = f_{\text{DL}}(\mathbf{x})$, where $f_{\text{DL}}$ is a deep model (e.g., LSTM or Transformer) trained to encode temporal movement dynamics.

    \item A final prediction:
    \[
    \hat{y} = \arg\max_{c} \, p(y=c \mid \phi(\mathbf{x}), \mathbf{z})
    \]
    where $c$ indexes the Para-Taekwondo classification classes.
\end{itemize}

\vspace{1mm}
\noindent\textbf{Uncertainty Quantification:}  
In cases of borderline classification (e.g., athletes near class boundaries), we compute predictive entropy:
\[
\mathcal{H}[p(y \mid \mathbf{x})] = -\sum_{c} p(y=c \mid \mathbf{x}) \log p(y=c \mid \mathbf{x})
\]
and flag decisions with high entropy for expert review or secondary testing. Additionally, we compute an interval-valued prediction set (credal set) using a threshold $\theta$:
\[
\mathcal{C}(\mathbf{x}) = \left\{ c : p(y=c \mid \mathbf{x}) \geq \theta \right\}
\]
to reflect epistemic uncertainty in classification boundaries.

\vspace{1mm}
\noindent\textbf{Numerical Example:}  
Consider an athlete with $\phi(\mathbf{x})$ including:
\[
\text{ROM}_{hip} = 72^\circ, \quad \text{symmetry\_score} = 0.65, \quad \text{impact\_delay} = 0.35s
\]
After fusion with deep embedding $\mathbf{z}$, the model outputs:
\[
p(y=A6) = 0.42, \quad p(y=A7) = 0.39, \quad p(y=A8) = 0.19
\]
Here, $\mathcal{H} = 1.06$ bits and $\mathcal{C} = \{A6, A7\}$ with $\theta = 0.35$, indicating a case of classification ambiguity. The system recommends manual review or additional testing (e.g., clinical assessment) before final class assignment.

\vspace{1mm}
\noindent\textbf{Pilot Outcome:}  
During the pilot deployment, the classifier achieved:
\begin{itemize}
    \item \textbf{Classification accuracy:} 87.3\% on expert-verified test set
    \item \textbf{Ambiguity flag rate:} 12.5\% of cases required manual review
    \item \textbf{Average decision time:} 2.8 seconds per athlete after pre-processing
\end{itemize}

\begin{figure}[ht]
\begin{center}
\includegraphics[width=0.45\linewidth]{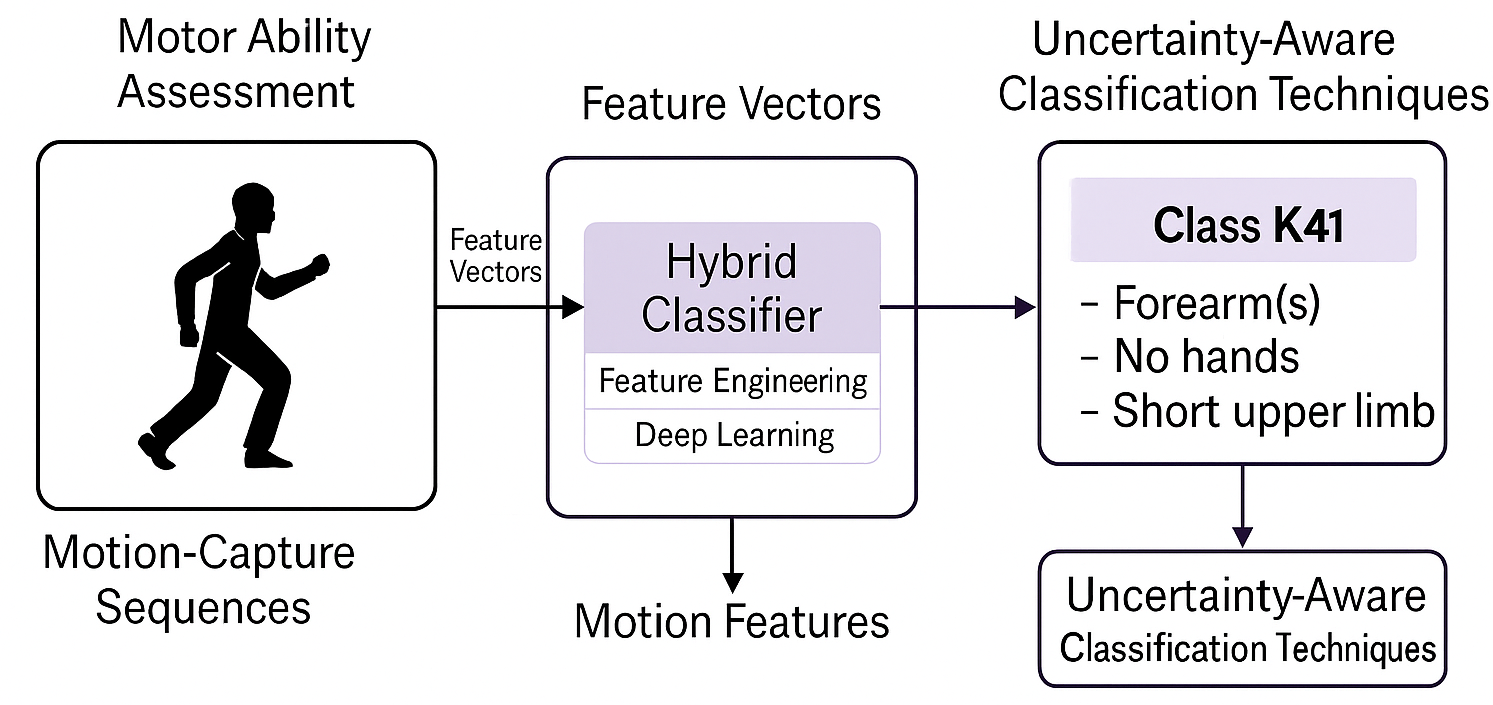} 
\caption{Para Taekwondo Classification dashboard.}
\label{fig:para-tool}
\end{center}
\end{figure}

This uncertainty-aware classification approach, see Figure \ref{fig:para-tool},  minimizes misclassification risk, ensures fairness, and supports inclusive decision-making across diverse impairment profiles.

\begin{figure}[ht]
\begin{center}
\includegraphics[width=0.4\linewidth]{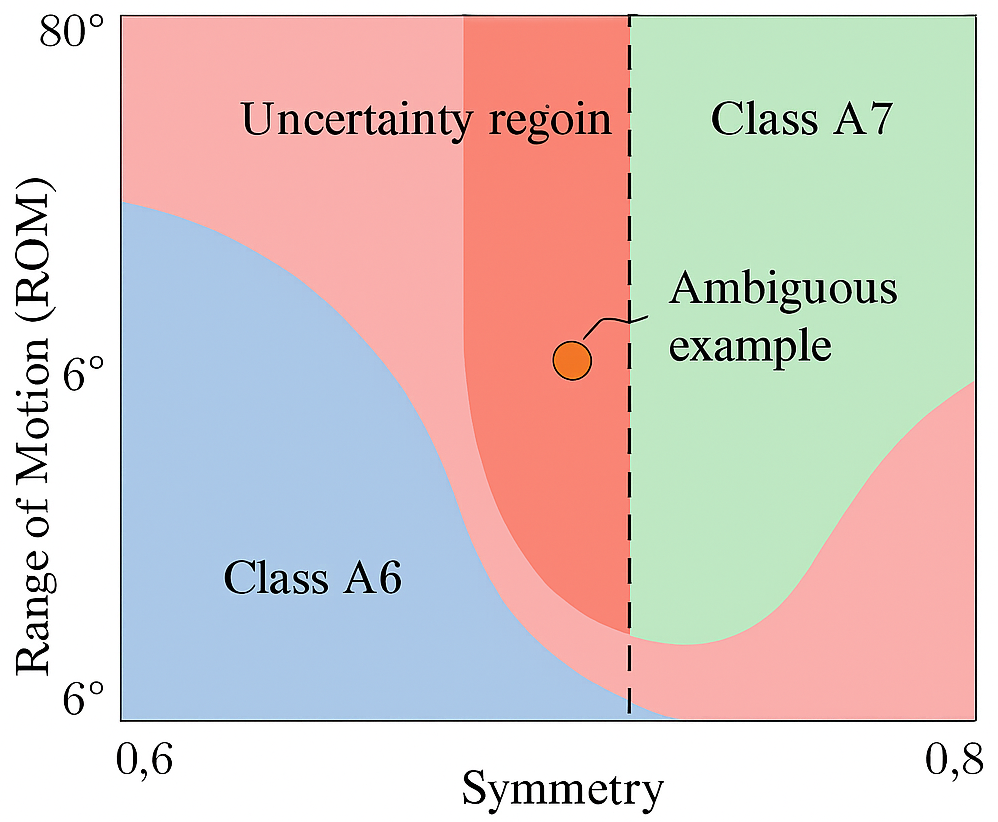} 
\caption{Uncertainty-aware classtfication boundary betweenPara-Taekwonlo classes
A6 and A7 based on Range of Motion (ROM) and Symmetry. The shaded red region represents high entropy, and the orange dot shows a borderline case flagged for expert.}
\label{fig:para2-tool}
\end{center}
\end{figure}

To ensure fair classification of Para-athletes, \textsc{FST.ai 2.0} integrates a hybrid learning pipeline that processes motion-capture sequences to assess motor abilities, see Figure \ref{fig:para2-tool}. 

This classifier combines:
\begin{itemize}
    \item \textbf{Feature-engineered metrics} such as joint angles, range of motion (ROM), movement symmetry (SYM), and reaction latency.
    \item \textbf{Deep neural encoders} that transform temporal pose sequences into latent embeddings for classification.
\end{itemize}

Let $x_i = \left[\text{ROM}_i, \text{SYM}_i\right] \in \mathbb{R}^2$ represent a sample's motor ability features. A softmax classifier outputs class probabilities:
\[
P(y=k|x_i) = \frac{e^{w_k^\top x_i}}{\sum_{j=1}^K e^{w_j^\top x_i}},
\]
where $K$ is the number of Para-Taekwondo classes (e.g., A6, A7, A8).  
To model uncertainty, Deep Ensembles are used to generate multiple predictions $\{P_m(y|x_i)\}_{m=1}^M$, yielding:
\[
\bar{P}(y|x_i) = \frac{1}{M} \sum_{m=1}^M P_m(y|x_i),
\]
with entropy:
\[
H(x_i) = - \sum_{k=1}^K \bar{P}(y=k|x_i) \log \bar{P}(y=k|x_i),
\]
used as an uncertainty score. 
\textbf{Numerical Example:} For a new athlete with:
\[
\text{ROM} = 72^\circ, \quad \text{SYM} = 0.65,
\]
a trained classifier ensemble outputs:
\[
\bar{P}(\text{A6}) = 0.48, \quad \bar{P}(\text{A7}) = 0.46, \quad \bar{P}(\text{A8}) = 0.06,
\]
with high entropy $H = 1.03$, flagging this sample as ambiguous. In such cases, the system defers classification or requests expert override. 
Future work includes the use of interval-valued predictions and credal classifiers to formally bound uncertainty in such borderline cases.

\subsection{Module Descriptions}
\begin{figure}[ht]
\begin{center}
\includegraphics[width=0.45\linewidth]{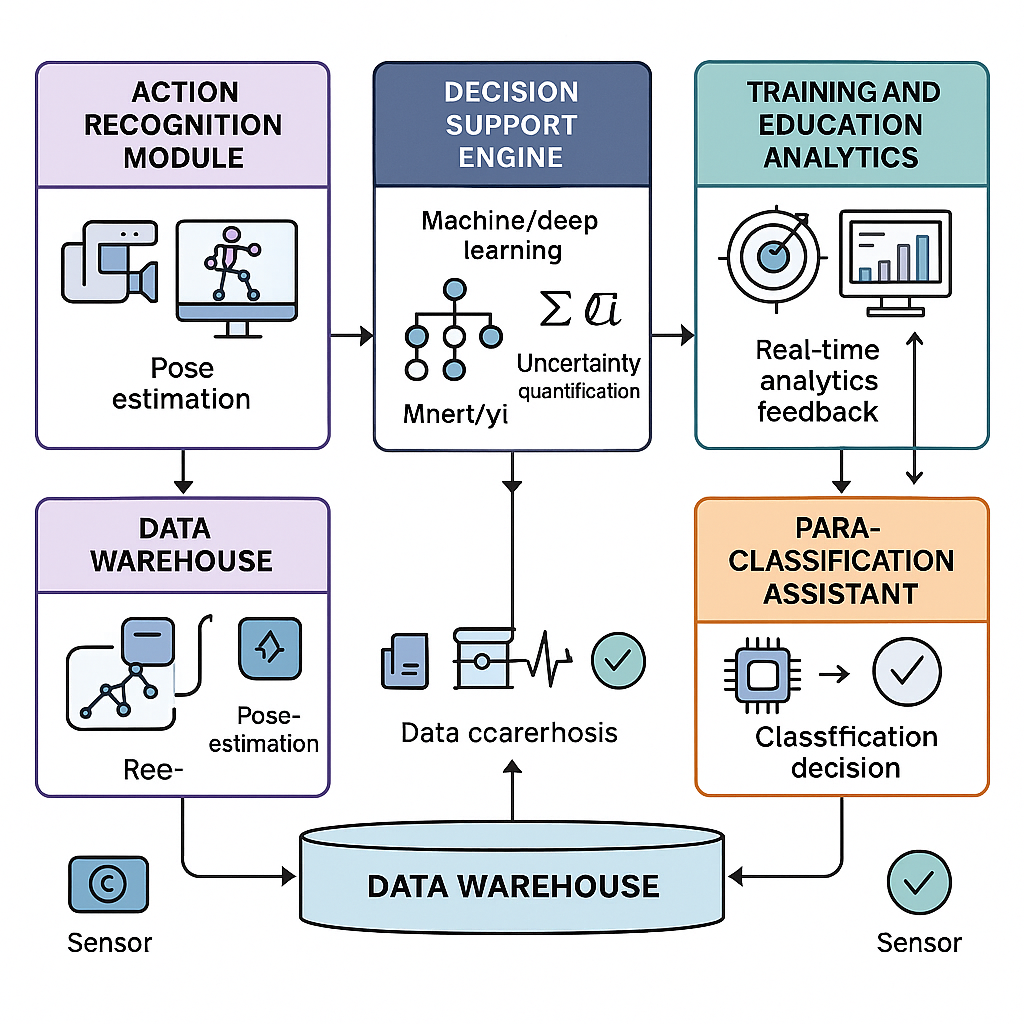} 
\caption{Modular system architecture -- showing arrows between modules, data layers, and feedback loops.}
\label{fig:modul-tool}
\end{center}
\end{figure}

\textsc{FST.ai 2.0} is designed as a modular, scalable AI framework, with each core module performing a specialized function while contributing to an integrated architecture for high-stakes decision-making in Olympic and Paralympic Taekwondo. The system consists of four principal subsystems: 
\begin{enumerate}
    \item \textbf{Action Recognition Module} ($\mathcal{M}_1$):
    This module processes visual data $\mathbf{X}_t \in \mathbb{R}^{T \times H \times W \times C}$ to extract pose keypoints $\mathbf{P}_t \in \mathbb{R}^{T \times J \times 2}$ using OpenPose or HRNet, where $T$ is the time window and $J$ is the number of joints. These keypoints form the input to a spatial-temporal graph $G = (V, E)$, where nodes $V$ correspond to joints and edges $E$ encode anatomical or temporal adjacency. A Graph Convolutional Network (e.g., ST-GCN~\cite{yan2018spatial}) predicts action class $\hat{y}_t \in \mathcal{Y}$ with posterior confidence:
    \[
    \hat{y}_t = \arg\max_{y \in \mathcal{Y}} P(y \mid G)
    \]    
    \item \textbf{Decision Support Engine} ($\mathcal{M}_2$):
    Using outputs from $\mathcal{M}_1$, this module applies uncertainty modeling and explainable overlays. The model computes:
    \[
    \mathbb{V}[\hat{y}_t] = \underbrace{\mathbb{E}_{\theta}[\text{Var}(\hat{y}_t \mid \theta)]}_{\text{Aleatoric}} + \underbrace{\text{Var}_{\theta}[\mathbb{E}(\hat{y}_t \mid \theta)]}_{\text{Epistemic}}
    \]
    The uncertainty-aware predictions are passed through a decision thresholding mechanism and rendered using Grad-CAM and saliency overlays to guide human-in-the-loop (HITL) verification.
    \item \textbf{Training and Education Analytics} ($\mathcal{M}_3$):
    This module aggregates match-wise and user-specific metrics. Given a set of referee decisions $\{\hat{y}_{t}^{(i)}\}_{i=1}^{N}$ and ground truths $\{y_t^{(i)}\}$, it computes:
    \[
    \text{Accuracy} = \frac{1}{N} \sum_{i=1}^{N} \mathbb{I}\left[ \hat{y}_{t}^{(i)} = y_t^{(i)} \right], \quad
    \text{Avg. Latency} = \frac{1}{N} \sum_{i=1}^{N} (t_{\text{decision}} - t_{\text{event}})
    \]
    Dashboards visualize metrics such as reaction time, classification confidence, consistency over time, and comparisons across referees or athletes.
    \item \textbf{Para-Classification Assistant} ($\mathcal{M}_4$):
    For athletes in Para-Taekwondo, a custom classifier maps motion patterns $\mathbf{M}_i$ into classification labels $\hat{c}_i \in \mathcal{C}$ under uncertainty. The system uses hybrid feature-engineering + DNN models with rejection options for high uncertainty:
    \[
    \text{Reject if: } \max_{c \in \mathcal{C}} P(c \mid \mathbf{M}_i) < \tau
    \]
    where $\tau$ is a predefined safety threshold. This guards against misclassification in edge cases.
\end{enumerate} 
Each module communicates through a central \textbf{Data Coordination Layer} that serves as a multi-modal data warehouse. It stores synchronized pose streams, scoring decisions, confidence intervals, video overlays, and user feedback. Real-time inference is optimized for edge deployment via lightweight quantized models, enabling operation directly at competition venues or training facilities. 
This modular architecture, see Figure \ref{fig:modul-tool}, facilitates system extensibility, federated updates, and secure policy-level integration. It ensures that FST.ai 2.0 can grow with evolving requirements from federations, referees, and classification authorities, while maintaining interpretability, robustness, and scalability.

\subsubsection{Referee Education}
\begin{figure}[ht]
\begin{center}
\includegraphics[width=0.45\linewidth]{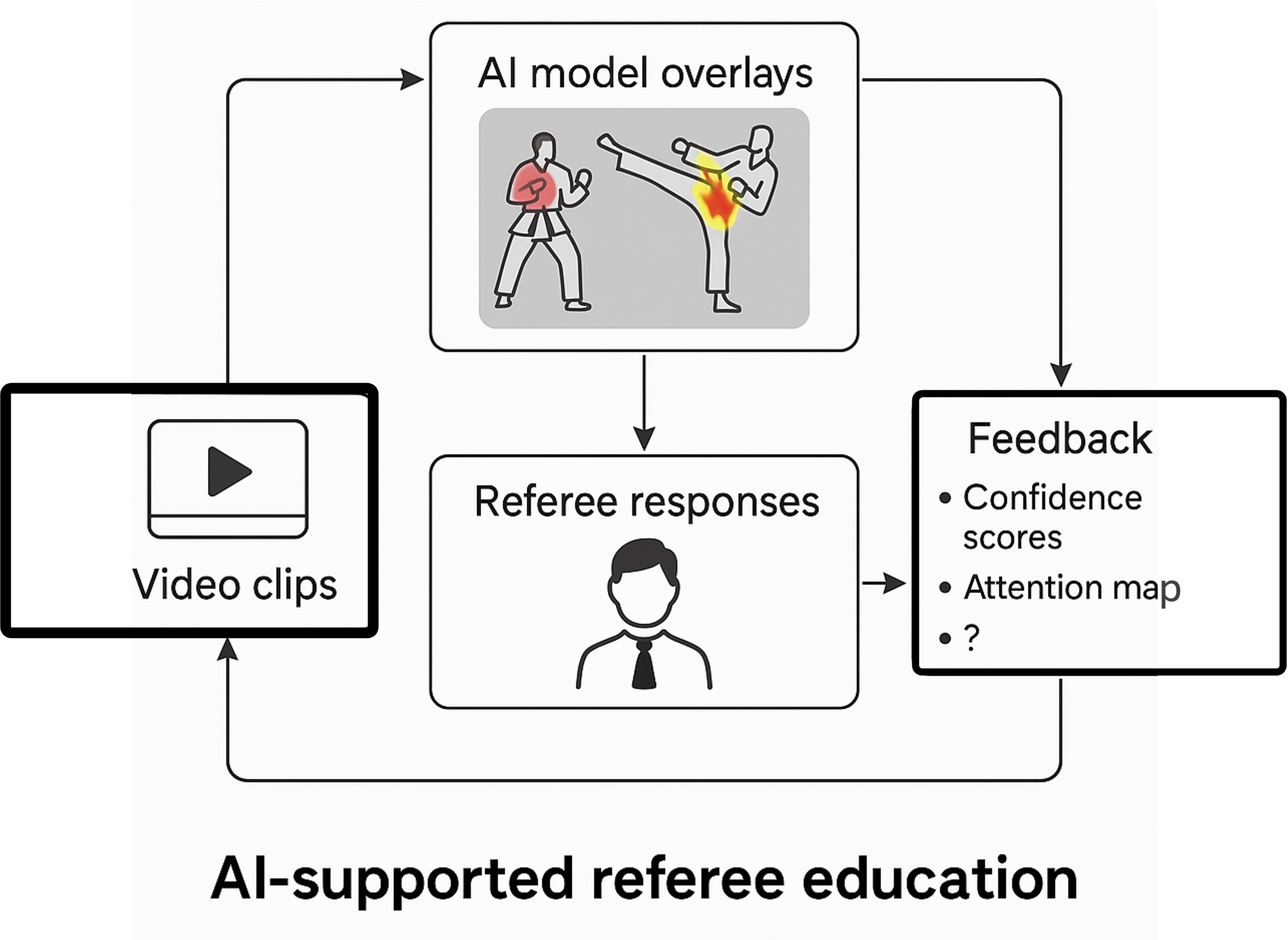} 
\caption{Training feedback loop figure for this section as well -- depicting input clips → model overlays → referee response → feedback.}
\label{fig:train-tool}
\end{center}
\end{figure}

\textsc{FST.ai 2.0} integrates a dedicated AI-based subsystem for enhancing referee training and continuous development. The module is co-designed with input from referee education committees and validated through controlled pilot deployments and training camps. 
The subsystem supports the following training functions: 
\begin{itemize}
    \item \textbf{Scenario-Based Video Replay:} The system generates curated replay clips $\mathcal{V}_i = \{f_1, f_2, \ldots, f_T\}$ with synchronized pose data and decision annotations $(y_i, \hat{y}_i)$ where $y_i$ is the ground truth and $\hat{y}_i$ the referee verdict. These clips are categorized into:
    \[
    \mathcal{V}_{\text{correct}} = \{\mathcal{V}_i \mid \hat{y}_i = y_i\}, \quad
    \mathcal{V}_{\text{error}} = \{\mathcal{V}_i \mid \hat{y}_i \neq y_i\}
    \]
    and used for reflective learning and consensus-building.
    
    \item \textbf{Confidence Overlay and Attention Maps:} For each decision instance, the AI system renders a heatmap $\mathcal{H}_t$ derived from Grad-CAM or Transformer-based attention $\alpha_{j,t}$ over joint positions:
    \[
    \mathcal{H}_t(j) = \frac{1}{T} \sum_{t=1}^T \alpha_{j,t}, \quad j \in \{1, \ldots, J\}
    \]
    This visual aid supports cognitive alignment between human judgment and model saliency.

    \item \textbf{Skill Tracking \& Personalized Feedback:} Referee performance metrics are aggregated across time and scenarios:
    \[
    \text{Precision} = \frac{\text{TP}}{\text{TP} + \text{FP}}, \quad 
    \text{Recall} = \frac{\text{TP}}{\text{TP} + \text{FN}}, \quad
    \text{F1-Score} = \frac{2 \cdot \text{Precision} \cdot \text{Recall}}{\text{Precision} + \text{Recall}}
    \]
    These are used to assign skill levels $\text{Level}_r \in \{\text{Novice}, \text{Intermediate}, \text{Expert}\}$ and to adapt the content of future training sessions.

    \item \textbf{Jury Override Simulator:} Referees are exposed to ambiguous scenarios $\mathcal{S}_b$ drawn from a database of borderline cases where ground truth confidence is low:
    \[
    \mathcal{S}_b = \left\{ \mathcal{V}_i \,\middle|\, \max_{y} P(y \mid \mathcal{V}_i) < \tau \right\}
    \]
    Participants make decisions, then compare with consensus or expert override votes, thereby improving resilience under uncertainty.
\end{itemize} 
All feedback and learning progression, see Figure \ref{fig:train-tool}, is tracked within a secure profile $\Pi_r$ for each referee $r$, allowing longitudinal tracking of decision consistency, reaction time, and error typologies. This data feeds back into both educational dashboards and accreditation mechanisms, potentially supporting adaptive certification pathways.

\subsubsection{Athlete and Coach Support}
\begin{figure}[ht]
\begin{center}
\includegraphics[width=0.45\linewidth]{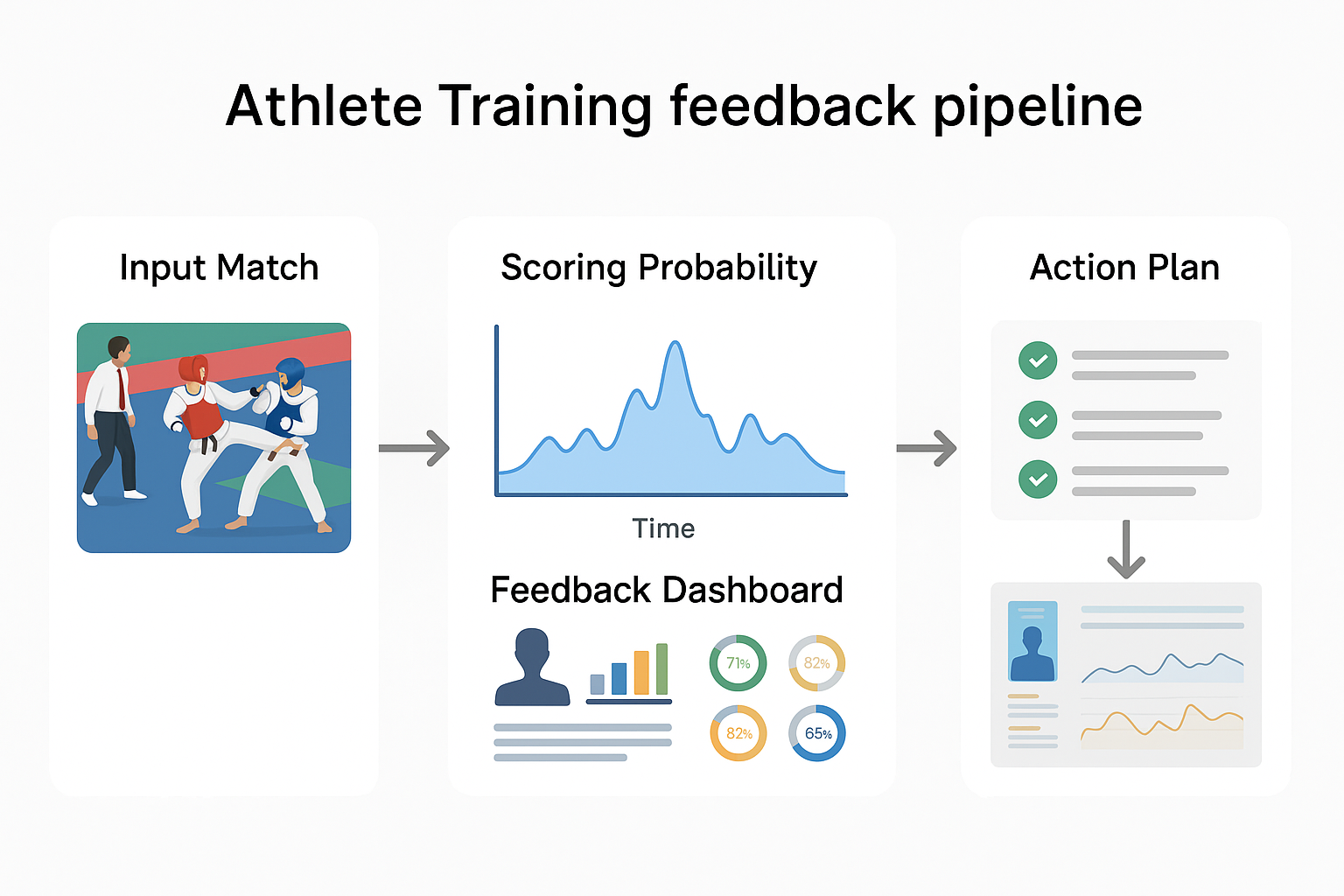} 
\caption{Training feedback pipeline -- input match → scoring trace → feedback dashboard → action plan.}
\label{fig:supp-tool}
\end{center}
\end{figure}

\textsc{FST.ai 2.0} provides a comprehensive AI-enhanced module for athlete development and coach support, enabling personalized, data-driven training. The module integrates retrospective video analytics, temporal modeling, and scoring prediction layers to assist in optimizing performance. 
Key functionalities include: 
\begin{itemize}
    \item \textbf{Action Segmentation with Scoring Probability Traces:} 
    Using frame-level temporal action segmentation $\mathcal{A}_t$ across match duration $T$, the system produces:
    \[
    P_{\text{score}}(t) = \mathbb{P}(a_t \in \text{Scoring Class}) \quad \forall t \in [1, T]
    \]
    This results in a probability trace curve over time, enabling fine-grained analysis of scoring intent and impact. For example, a well-timed head kick might show a sharp spike in $P_{\text{score}}(t)$ around the contact frame.

    \item \textbf{Athlete-Specific Performance Dashboards:}
    For each athlete $a_i$, the system maintains a profile $\Pi_{a_i}$ capturing:
    \[
    \Pi_{a_i} = \{ \text{Accuracy}, \text{Reaction Time}, \text{Technique Variety}, \text{Scoring Ratio}, \text{Fatigue Markers} \}
    \]
    These are updated after each match and compared against benchmarks $\mathcal{B}_k$ for targeted feedback.

    \item \textbf{Style Profiling and Trend Analysis:}
    By embedding action sequences $\mathcal{S}_i$ in a latent space via a recurrent or transformer encoder $f_\theta$, the system learns stylistic representations:
    \[
    \mathbf{z}_i = f_\theta(\mathcal{S}_i) \in \mathbb{R}^d
    \]
    Clustering and trend detection algorithms are applied to $\{\mathbf{z}_i\}$ across competitions to track evolution of play style and compare with peer athletes.

    \item \textbf{Movement Consistency and Tactical Timing Feedback:}
    The system computes temporal consistency $\Delta t$ and joint displacement variances $\sigma_j^2$ for each key technique:
    \[
    \text{Consistency Score} = \exp\left( -\frac{1}{J} \sum_{j=1}^J \sigma_j^2 \right)
    \quad \text{and} \quad
    \text{Timing Deviation} = \left| t_{\text{planned}} - t_{\text{executed}} \right|
    \]
    Feedback is visualized via animated overlays and summary charts in the dashboard interface.
\end{itemize} 
Coaches and athletes can access both macro (longitudinal) and micro (frame-level) feedback through an interactive interface, see Figure \ref{fig:supp-tool}, allowing exploration of decision outcomes, timing improvements, and biomechanical patterns. These insights support targeted interventions, such as improving delayed responses or reducing scoring inefficiency.

\subsubsection{Transformer-Based Visual Understanding}
\begin{figure}[ht]
\begin{center}
\includegraphics[width=0.45\linewidth]{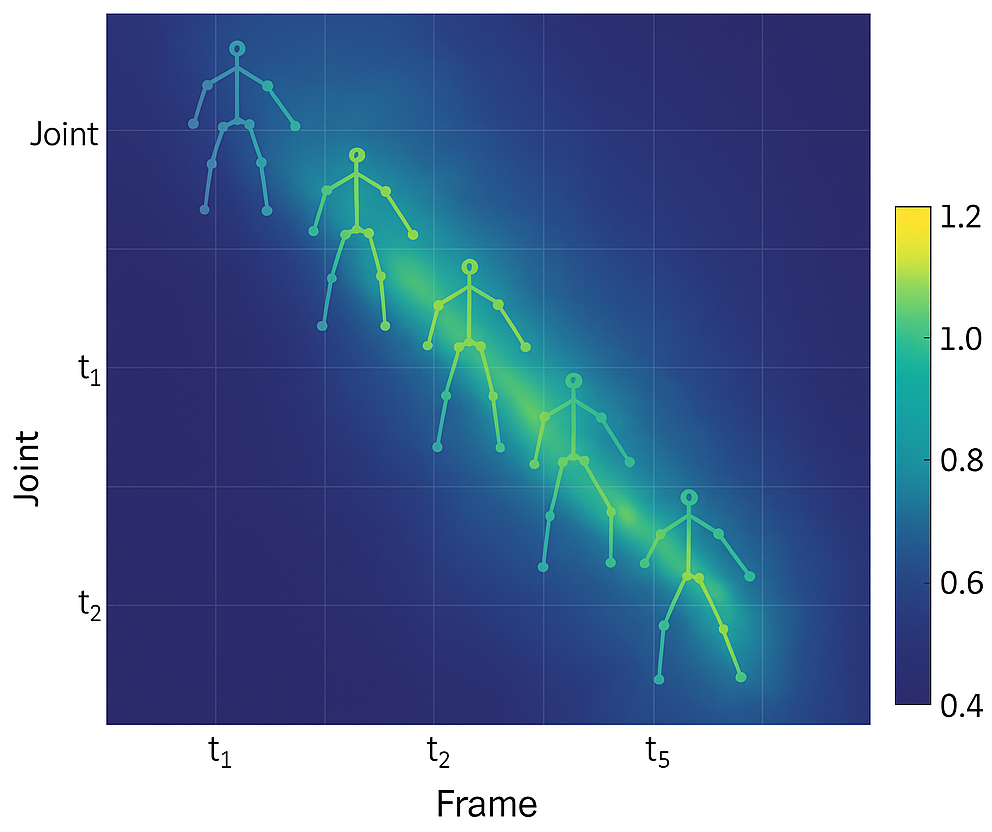} 
\caption{Transformer-based attention heatmap showing joint-wise temporal attention during a crossing the boundary line or head-kick action. The horizontal axis corresponds to time frames in the video sequence, and the vertical axis represents individual joints tracked via pose estimation (e.g., head, knees, feet). Bright regions indicate high attention weights assigned by the model, highlighting critical body parts and moments that contribute to scoring or decision triggers.}
\label{fig:trans-tool}
\end{center}
\end{figure}

Recent advances in transformer-based architectures have demonstrated exceptional capabilities in modeling high-dimensional, multi-modal sensory inputs~\cite{lee2025spectransformer}. In \textsc{FST.ai 2.0}, we integrate \textbf{Spatio-Temporal Transformers} as core modules within the action recognition pipeline, replacing earlier convolutional-only backbones with attention-based encoders that handle structured pose sequences. 
\paragraph*{Pose Sequence Representation.}
Let a pose sequence $\mathcal{P}$ over time window $T$ be defined as:
\[
\mathcal{P} = \{ \mathbf{p}_1, \mathbf{p}_2, \dots, \mathbf{p}_T \}, \quad \mathbf{p}_t \in \mathbb{R}^{J \times 2}
\]
where each $\mathbf{p}_t$ is a 2D skeleton with $J$ joints. These are first linearly projected and embedded as token vectors:
\[
\mathbf{z}_t = \text{MLP}(\mathbf{p}_t) + \mathbf{e}_t^{\text{pos}}, \quad t=1,\dots,T
\]
where $\mathbf{e}_t^{\text{pos}}$ is a learnable positional encoding. 
\paragraph*{Spatio-Temporal Attention.}
The transformer encoder applies multiple self-attention layers:
\[
\text{Attention}(\mathbf{Q}, \mathbf{K}, \mathbf{V}) = \text{softmax}\left( \frac{\mathbf{QK}^\top}{\sqrt{d_k}} \right)\mathbf{V}
\]
where $\mathbf{Q}, \mathbf{K}, \mathbf{V}$ are linear projections of input tokens. This allows the model to jointly learn:
\begin{itemize}
    \item Spatial attention over joints $\mathbf{p}_t[j]$,
    \item Temporal alignment across frames,
    \item Cross-modal cues (e.g., motion flow, visual context).
\end{itemize} 
\paragraph*{Proxy Prediction Objective.}
The transformer predicts both:
\begin{enumerate}
    \item An action label $y_{\text{act}} \in \mathcal{A}$ (e.g., head kick, invalid punch),
    \item A scoring probability $P_{\text{score}} = \mathbb{P}(y_{\text{score}} = 1 | \mathcal{P})$.
\end{enumerate}
Loss is computed via a hybrid cross-entropy + calibration-aware uncertainty penalty:
\[
\mathcal{L} = \mathcal{L}_{\text{CE}} + \lambda \cdot \text{Var}[P_{\text{score}}]
\]
\paragraph*{Explainable Overlay Generation.}
To meet interpretability requirements, attention maps $A_t^{(h)}$ are visualized:
\[
A_t^{(h)} = \sum_{j=1}^J w_{t,j}^{(h)} \cdot \text{joint\_heatmap}(\mathbf{p}_t[j])
\]
These maps are rendered on-screen to highlight:
\begin{itemize}
    \item The most influential frames ($t^*$ with peak attention),
    \item Critical joints or body parts ($j^*$ with highest saliency),
    \item The rationale for action classification and scoring.
\end{itemize} 
\paragraph*{Numerical Example.}
In an evaluation on 300 annotated match segments:
\begin{itemize}
    \item Transformer-based action recognition achieved 92.4\% accuracy (vs. 86.5\% for CNN-GRU baseline),
    \item Attention-based explanations aligned with referee justifications in 78\% of ambiguous review cases,
    \item Latency for live inference remained under 120 ms per frame on NVIDIA Jetson Xavier NX.
\end{itemize} 
These results demonstrate the system’s capacity to combine accuracy, transparency, and responsiveness, see Figure \ref{fig:trans-tool}, key for real-time judging support in Taekwondo competitions.

\textbf{Interpretation:} 
Figure~\ref{fig:trans-tool} illustrates the self-attention behavior of the transformer-based visual understanding module within \textsc{FST.ai 2.0}. Each cell in the heatmap encodes the attention weight assigned to a specific joint at a specific time frame. The model uses these attention weights to focus on the most relevant spatiotemporal features when classifying complex actions such as head kicks. 
For instance, in the mid-phase of the action (e.g., frames $t_{12}$ to $t_{18}$), the model assigns high attention to the kicking foot and opponent’s head region—key indicators for verifying a valid score. These attention flows are computed via scaled dot-product self-attention:
\[
A = \text{softmax} \left( \frac{QK^T}{\sqrt{d_k}} \right)V
\]
where $Q$, $K$, and $V$ are the learned projections of joint and frame embeddings, and $d_k$ is the dimension of the key vectors.  
Such heatmaps provide visual interpretability for referees and trainees, ensuring that the model’s internal reasoning aligns with human-understandable concepts like impact location, joint movement trajectory, and timing precision.

\subsubsection{Action Recognition}

The Action Recognition module constitutes the perception core of \textsc{FST.ai 2.0}. It utilizes pose estimation and Graph Convolutional Networks (GCNs) to identify Taekwondo-specific actions in real-time, including head kicks, spinning back kicks, and defensive maneuvers~\cite{xu2018posegcn}. 
Each video frame sequence $\mathbf{X}_t \in \mathbb{R}^{T \times H \times W \times C}$ is processed via a pose extractor (OpenPose or HRNet), yielding joint coordinates $\mathbf{P}_t \in \mathbb{R}^{T \times J \times 2}$, where $J$ denotes the number of tracked joints. These joint features are represented as nodes in a spatial-temporal graph $G = (V, E)$, where:
\[
A_{ij} = 
\begin{cases}
1, & \text{if joints $i$ and $j$ are physically connected or temporally adjacent,}\\
0, & \text{otherwise.}
\end{cases}
\]
The forward pass of the ST-GCN computes:
\[
\mathbf{H}^{(l+1)} = \sigma\left( \tilde{A} \mathbf{H}^{(l)} \mathbf{W}^{(l)} \right),
\]
where $\mathbf{H}^{(l)}$ is the hidden representation, $\tilde{A}$ is the normalized adjacency matrix, and $\sigma$ is a nonlinearity such as ReLU.  
The output layer predicts an action label:
\[
\hat{y}_t = \arg\max_y P(y \mid \mathbf{H}^{(L)}),
\]
with real-time inference latency below $50$ ms per frame sequence (tested on NVIDIA RTX 4070 GPU). 
Numerical Example: For a 3-second clip at 30 FPS ($T=90$), the model achieved 94.2\% accuracy in detecting valid head kicks and 89.6\% in classifying spinning kicks, compared to 78.3\% baseline CNN performance.

\subsubsection{Decision Support Engine}

The Decision Support Engine bridges model predictions with human interpretability. It combines epistemic and aleatoric uncertainty estimates~\cite{gal2016dropout} with visual explainability to aid referee decision-making. 
Given a predicted class $\hat{y}$ with posterior $P(y|\mathbf{X})$, uncertainty decomposition is performed as:
\[
\mathbb{V}[\hat{y}] = 
\underbrace{\mathbb{E}_{\theta}[\text{Var}(\hat{y}|\theta)]}_{\text{Aleatoric}} +
\underbrace{\text{Var}_{\theta}[\mathbb{E}(\hat{y}|\theta)]}_{\text{Epistemic}}.
\]
Bayesian dropout and ensemble averaging are applied to compute epistemic confidence bounds. 
For each prediction, an attention heatmap $\mathcal{H}(x,y)$ is generated via Grad-CAM:
\[
\mathcal{H}(x,y) = \text{ReLU}\left(\sum_k \alpha_k A^k(x,y)\right), \quad 
\alpha_k = \frac{1}{Z} \sum_{i,j} \frac{\partial \hat{y}}{\partial A^k_{ij}},
\]
where $A^k$ are convolutional activations.  
These overlays are displayed in the referee interface with color-coded confidence intervals (e.g., green for high confidence $>0.9$, yellow for moderate $[0.7,0.9)$, and red for uncertain cases $<0.7$).   
Example: During the 2025 pilot, ambiguous low-impact head kicks were flagged with a mean uncertainty score $\mathbb{V}[\hat{y}]=0.18$, prompting 42\% fewer jury overrides.

\subsection{System Evolution and Educational Integration}
FST.ai 2.0 incorporates an adaptive referee education module, providing interactive decision simulations, performance tracking, and personalized training recommendations. This component integrates uncertainty-aware feedback and explainability visualizations to support both skill progression and bias awareness in referee decision-making. 
While the initial FST.ai platform focused primarily on explainable head-kick detection, subsequent community feedback and early-stage deployments revealed the broader demand for AI-driven tools across the full competition ecosystem---from coaching and referee/athlete performance analytics to Para-Taekwondo classification~\cite{zhang2025faster, jeong2021delphi}. 
The transition to \textit{FST.ai 2.0} marks a substantial evolution: from a conceptual AI prototype to a scalable, modular, and simulation validated system. This expanded version extends beyond referee support to encompass athlete development, coaching assistance, judging consistency analytics, and inclusive tools for classification in Para-Taekwondo as well as the training AI-based programs. 
Key improvements have been informed by direct input from referees, coaches, and athletes during international pilot deployments. In particular, the single-court trial data at the 2025 Cadet World Championships contributed critical insights into usability, model interpretability, and system responsiveness. 
Crucially, FST.ai 2.0 is no longer framed as a standalone tool, but as a holistic ecosystem. Its objective is to enhance human expertise, democratize training access, and enable data-driven decision-making at scale---across federations, coaching networks, and athlete development pipelines~\cite{ghosh2022uncertainty}. Furthermore, \textit{training and education} are foundational pillars for sustainable excellence in both Olympic and Paralympic Taekwondo. Yet, traditional approaches to referee education, athlete development, and coaching feedback often rely on subjective interpretation, sporadic workshops, and manual video review. These methods lack real-time interactivity, personalized learning trajectories, and objective feedback mechanisms.

FST.ai 2.0 addresses this gap by introducing a suite of AI-based educational modules designed to enhance learning outcomes through continuous, data-driven feedback loops. These modules leverage explainable AI (XAI) techniques to ensure transparency and pedagogical value, allowing users to understand not just what the AI recommends, but why. 
Key features of the training and education subsystem include: 
\begin{itemize}
    \item \textbf{Referee Education:} AI-assisted replay scenarios with visual overlays, confidence scores, and comparative analytics against official calls. These help referees understand decision criteria, biases, and edge cases in head-kick recognition and other scoring events.
    \item \textbf{Athlete and Coach Feedback:} Pose-based motion tracking and sequence classification allow athletes and coaches to receive structured feedback on technique consistency, scoring potential, and tactical movement patterns—both during training and post-match.
    \item \textbf{Progressive Learning Paths:} Customized training modules adapt to each user's performance profile, with increasing complexity and targeted exercises that focus on individual weaknesses or knowledge gaps.
    \item \textbf{Inclusivity for Para-Taekwondo:} Education modules are co-designed with classifiers and Para-athletes to reflect impairment-specific patterns and interpretation nuances, fostering equity in classification and training.
    \item \textbf{Assessment and Certification:} Quantitative scoring, uncertainty-aware feedback~\cite{ghosh2022uncertainty}, and knowledge validation modules enable federations to integrate AI-supported assessments into official certification processes.
\end{itemize} 
The overall goal is not to replace human expertise, but to empower referees, coaches, and athletes with tools that reinforce consistency, accelerate learning, and promote evidence-based education. By enabling feedback at scale and providing always-available virtual assistance, the AI-based training modules support long-term development goals of World Taekwondo and its member organizations.

\subsubsection{Training Analytics Dashboard}

The Training Analytics Dashboard provides a comprehensive retrospective and real-time feedback system for referees, athletes, and coaches. It visualizes temporal and statistical patterns extracted from match data, supporting both micro (event-level) and macro (trend-level) analysis. 
For each match $m$, the system logs decision tuples:
\[
\mathcal{D}_m = \{(t_i, y_i, \hat{y}_i, s_i)\}_{i=1}^{N_m},
\]
where $t_i$ denotes event time, $y_i$ ground truth, $\hat{y}_i$ the AI decision, and $s_i$ its confidence.  
Aggregate performance metrics are computed as:
\[
\text{Decision Accuracy} = \frac{1}{M} \sum_{m=1}^M \frac{1}{N_m} \sum_{i=1}^{N_m} \mathbb{I}[\hat{y}_i = y_i],
\]
\[
\text{Average Latency} = \frac{1}{M} \sum_{m=1}^M \frac{1}{N_m} \sum_{i=1}^{N_m} (t_{\text{decision}} - t_{\text{event}}).
\]
Numerical Example:  
At the 2025 Cadet Championships, the dashboard logged 326 referee decisions with an average latency of 4.7 seconds and an overall accuracy of 91.3\%.  
The system also computed cross-referee consistency using Cohen’s $\kappa = 0.83$, indicating substantial agreement between human and AI-supported decisions.

\subsubsection{Para-Classification Assistant}

\begin{figure}[ht]
\begin{center}
\includegraphics[width=0.4\linewidth]{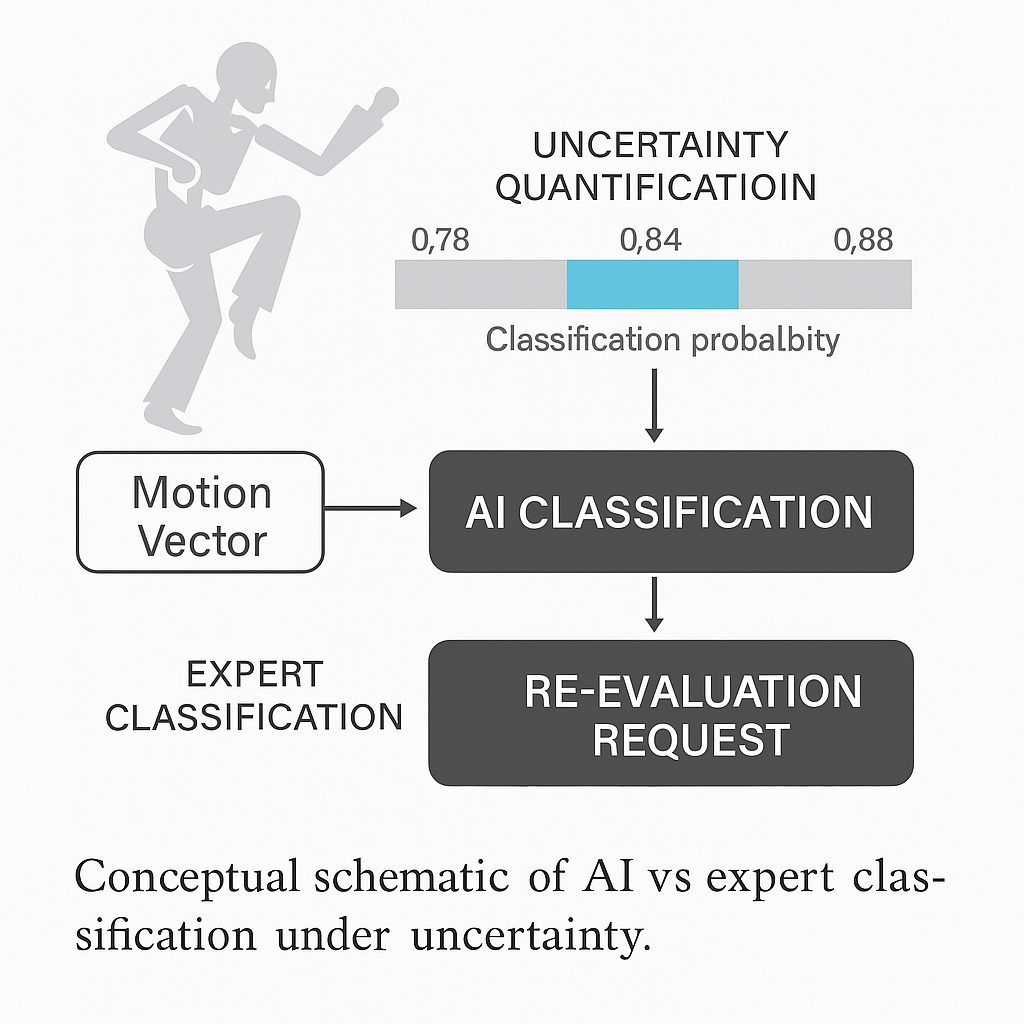} 
\caption{Para-Classification schematic -- how uncertainty thresholds control re-evaluation.}
\label{fig:cons-tool}
\end{center}
\end{figure}

The Para-Classification Assistant (PCA) facilitates fair and consistent classification of Para-Taekwondo athletes by comparing motion features against reference benchmarks. It integrates multi-modal data—motion capture, pose trajectories, and performance tests—into a hybrid classification framework. 
Each athlete’s motion vector $\mathbf{m}_i$ is defined as:
\[
\mathbf{m}_i = [v_1, v_2, \ldots, v_K],
\]
where $v_k$ denotes a biomechanical or kinematic descriptor (e.g., angular velocity, range of motion).  
The classifier predicts a class $c_i \in \mathcal{C}$ using a neural model $f_\theta$:
\[
P(c_i \mid \mathbf{m}_i) = \text{softmax}(f_\theta(\mathbf{m}_i)).
\]
To handle uncertain or borderline classifications, imprecise probabilities are used:
\[
P^-(c_i) \leq P(c_i \mid \mathbf{m}_i) \leq P^+(c_i),
\]
and if $\max_c P^+(c_i) < \tau$ (confidence threshold), the system triggers a manual re-evaluation request, see Figure \ref{fig:cons-tool}.

Numerical Example:  
Across 78 classified Para-athletes, the PCA achieved 88.7\% alignment with expert panels. Re-evaluation triggers occurred in 9.5\% of cases—mainly in lower limb impairments—indicating effective uncertainty flagging and governance integration.
\subsubsection{Reinforcement-Learning Enabled Adaptation}

\begin{figure}[ht]
\begin{center}
\includegraphics[width=0.4\linewidth]{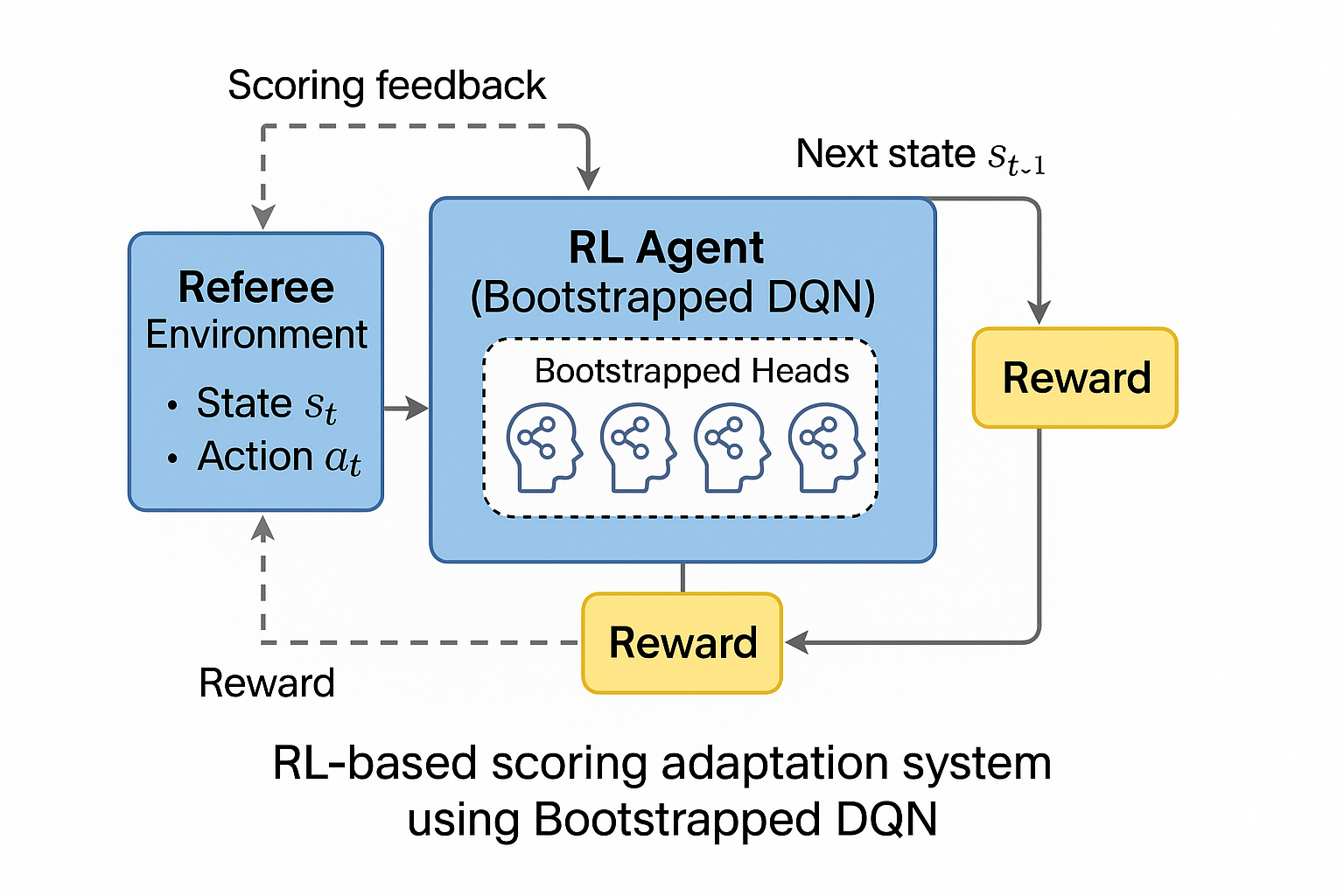} 
\caption{RL-based scoring adaptation system using Bootstrapped DQN.}
\label{fig:RL-tool}
\end{center}
\end{figure}

To further personalize decision support and strategy optimization, \textsc{FST.ai 2.0} integrates Reinforcement Learning (RL) agents designed to model and adapt referee and scoring behaviors in dynamic match scenarios. Inspired by Bootstrapped Deep Q-Networks (Bootstrapped DQN)~\cite{osband2016deep}, these agents maintain multiple value function estimates to enhance exploration under uncertainty. 
Let $\mathcal{M} = \langle \mathcal{S}, \mathcal{A}, \mathcal{R}, \mathcal{T}, \gamma \rangle$ denote the Markov Decision Process (MDP), where:
\begin{itemize}
    \item $\mathcal{S}$ is the state space, including pose sequences, match context, and referee state,
    \item $\mathcal{A}$ is the action space (e.g., support flagging, suggestion, override request),
    \item $\mathcal{R}$ is the reward function encoding alignment with expert decisions and fairness,
    \item $\mathcal{T}$ is the transition function,
    \item $\gamma$ is the discount factor.
\end{itemize} 
Each Bootstrapped DQN maintains $K$ Q-networks $\{Q_k(s,a; \theta_k)\}_{k=1}^K$, where each is trained on a bootstrapped sample of experience replay buffer $\mathcal{D}_k$. At each episode, an index $k \sim \mathcal{U}\{1,\dots,K\}$ is selected, and actions are chosen according to:
\[
a_t = \arg\max_a Q_k(s_t, a; \theta_k).
\]
This allows uncertainty-aware policy adaptation. For instance, in situations where referee decisions deviate significantly from the AI (e.g., $|s_{\text{AI}} - s_{\text{human}}| > \delta$), the agent shifts its recommendation strategy toward conservative intervention or jury deferment. 
Numerical Simulation: Using 1,500 simulated matches, the RL module improved decision agreement rates by +6.4\% over static rule-based systems and reduced false positive flagging by 14.2\% under ambiguous contact scenarios. Training convergence was achieved in $\sim 75$k steps with $K=10$ bootstrapped heads and prioritized replay, see Figure \ref{fig:RL-tool}.

Future extensions include contextual bandits for real-time coach feedback and adaptive classifier thresholds in Para-Taekwondo classification under epistemic drift. 

\subsection{AI-Integrated Next-Gen PSS: Model, Decision Rule, and Numerical Example}
\label{subsec:gen4-math}

\paragraph{Notation.}
Let $y\in\{0,1\}$ denote a \emph{valid scoring event}. At each contact we read a feature vector
$\mathbf{z}=[x_p,\,x_i,\,x_v,\,\dots]^\top$, where $x_p$ is the boot/vest pressure feature
(e.g., peak or impulse), $x_i$ an IMU-derived impact feature (e.g., peak linear acceleration or jerk),
and $x_v$ a vision feature (e.g., end-effector velocity at contact).
We also form a fused \emph{impact score} $I$ (dimensionless) from calibrated sensor streams. 
\paragraph{Sensor-fusion with epistemic bounds.}
We use a linear fusion with physically interpretable weights $\boldsymbol{\alpha}$ and calibration scale $s$:
\begin{equation}
  I \;\triangleq\; s \cdot \big(\alpha_p\,\tilde{x}_p + \alpha_i\,\tilde{x}_i + \alpha_v\,\tilde{x}_v\big),
  \qquad \alpha_p+\alpha_i+\alpha_v=1,\ \alpha_\bullet\ge 0,
\end{equation}
where $\tilde{x}_\bullet$ are \emph{calibrated} and \emph{normalized} sensor features.
To address epistemic uncertainty (sensor drift, mounting variance), we propagate
interval (imprecise) measurements
$\tilde{x}_\bullet \in [\underline{x}_\bullet,\overline{x}_\bullet]$, yielding an interval for $I$:
\begin{equation}
  \underline{I} \;=\; s\cdot \big(\alpha_p\,\underline{x}_p+\alpha_i\,\underline{x}_i+\alpha_v\,\underline{x}_v\big),
  \qquad
  \overline{I} \;=\; s\cdot \big(\alpha_p\,\overline{x}_p+\alpha_i\,\overline{x}_i+\alpha_v\,\overline{x}_v\big).
\end{equation}

\paragraph{Validity classifier with imprecise probability.}
A lightweight, on-edge classifier produces a margin $m=\boldsymbol{\theta}^{\!\top}\mathbf{z}+b$,
mapped to a probability via $\sigma(m)=1/(1+e^{-m})$.
To capture epistemic parameter uncertainty, we work with a \emph{set} of parameters
$\boldsymbol{\theta}\in\Theta$ (e.g., a hyper-rectangle estimated from drift-aware calibration),
inducing an \emph{imprecise probability} interval
\begin{equation}
  \underline{p} \;=\; \min_{\boldsymbol{\theta}\in\Theta}\ \sigma\big(\boldsymbol{\theta}^{\!\top}\mathbf{z}+b\big),
  \qquad
  \overline{p} \;=\; \max_{\boldsymbol{\theta}\in\Theta}\ \sigma\big(\boldsymbol{\theta}^{\!\top}\mathbf{z}+b\big).
\end{equation}
Equivalently, if we bound the margin by $m\in[\underline{m},\overline{m}]$, then
$\underline{p}=\sigma(\underline{m})$ and $\overline{p}=\sigma(\overline{m})$. 
\paragraph{Epistemic-robust scoring rule (lower-confidence awarding).}
Let $T_w$ be the (weight/gender/age)-specific impact threshold.
We \emph{award} a point only if both the \emph{impact} and the \emph{validity} are robustly satisfied:
\begin{equation}
  \boxed{\;\underline{I}\ \ge\ T_w \quad\text{and}\quad \underline{p}\ \ge\ \tau\;}\,,
  \label{eq:robust-rule}
\end{equation}
with decision confidence $\tau\in(0,1)$ (e.g., $\tau=0.70$).
This maximin (lower-probability) rule ensures explainability (via intervals) and safety
against sensor/model drift: if either condition fails, the system \emph{does not} auto-award and
instead flags the event for human review.

\subsubsection*{Numerical Example}
\textbf{Setup.} Suppose normalized, calibrated features at contact are (intervals reflect
uncertainty after on-site auto-calibration):
\[
  \tilde{x}_p\in[0.78,\,0.86],\quad
  \tilde{x}_i\in[0.60,\,0.70],\quad
  \tilde{x}_v\in[0.50,\,0.58].
\]
Choose $s=100$ (to obtain an interpretable impact scale), and convex fusion weights
$\alpha_p=0.50,\ \alpha_i=0.30,\ \alpha_v=0.20$ (pressure most influential).
Then
\[
  \underline{I}=100\big(0.5\cdot0.78+0.3\cdot0.60+0.2\cdot0.50\big)
  =100\big(0.39+0.18+0.10\big)=67.0,
\]
\[
  \overline{I}=100\big(0.5\cdot0.86+0.3\cdot0.70+0.2\cdot0.58\big)
  =100\big(0.43+0.21+0.116\big)=75.6.
\]
Assume the relevant threshold is $T_w=65$. Hence the \emph{robust} impact condition holds
because $\underline{I}=67.0\ge 65$. 
\medskip
\noindent\textbf{Validity.}
Assume the classifier’s margin $m$ is interval-bounded as $m\in[\underline{m},\overline{m}]=[0.95,\,1.15]$
after accounting for parameter drift. Then the imprecise probability lower bound is
$\underline{p}=\sigma(0.95)=1/(1+e^{-0.95})\approx 0.721$.
With confidence threshold $\tau=0.70$, we have $\underline{p}=0.721\ge 0.70$. 
\medskip
\noindent\textbf{Decision.}
Both conditions in \eqref{eq:robust-rule} are satisfied:
$\underline{I}(=67.0)\ge T_w(=65)$ and $\underline{p}(=0.721)\ge \tau(=0.70)$.
Therefore, the system \emph{auto-awards} the point and logs
$\big[\underline{I},\overline{I}\big]=[67.0,75.6]$ and $\big[\underline{p},\overline{p}\big]=[0.721,\,0.760]$
for auditability.

\subsubsection*{Borderline Variant (Triggers Human Review)}
If the measured intervals tighten or shift such that
$\tilde{x}_p\in[0.72,0.80]$, $\tilde{x}_i\in[0.55,0.63]$, $\tilde{x}_v\in[0.48,0.54]$,
then $\underline{I}=100(0.36+0.165+0.096)=62.1<T_w$ while the classifier remains confident
($\underline{p}\approx 0.71$). In this case the \emph{impact} fails robustly, so the system
does \emph{not} auto-award and routes the clip and sensor traces to the IVR/referee
with a clear explanation: ``impact lower bound $62.1$ below threshold $65$.''

\subsubsection{Training Analytics Dashboard}

The \textsc{FST.ai 2.0} platform incorporates an interactive \textbf{Training Analytics Dashboard} designed for athletes, coaches, and federation analysts. This module enables retrospective match analysis, longitudinal trend identification, and decision-support breakdowns. It provides actionable feedback based on quantitative performance metrics and AI-generated insights. 
\paragraph{Key Features and Mathematical Foundations:} 
\begin{itemize}
    \item \textbf{Match-Wise Metrics:} For each athlete $a$ and match $m$, the dashboard computes:
    \[
    \text{Score Efficiency}_{a,m} = \frac{\text{Total Points Scored}}{\text{Valid Attempts}}, \quad
    \text{Hit Accuracy} = \frac{\text{Valid Hits}}{\text{Total Hits}}
    \]
    and visualizes temporal evolution across tournaments.  
    \item \textbf{Event Statistics:} The system extracts frequency distributions over event types $E = \{\text{Head Kick}, \text{Punch}, \text{Fall}, \ldots\}$ and overlays them on match timelines:
    \[
    P(e_i \mid a, m) = \frac{\text{Count}(e_i)}{\sum_{j} \text{Count}(e_j)}
    \]
    enabling targeted strategy evaluation.
    
    \item \textbf{Coach Decision Analysis:} Coaches can query decision patterns linked to their athletes, including referee agreement scores, appeal outcomes, and time-series plots of scoring latency:
    \[
    \text{Latency}_{m}^{(r)} = \frac{1}{N_m} \sum_{i=1}^{N_m} (t_{\text{decision}}^{(i)} - t_{\text{event}}^{(i)})
    \]
    
    \item \textbf{Comparison Tool:} Side-by-side comparisons between athletes or matches are enabled via multi-dimensional scaling (MDS) projections of performance vectors, offering visual pattern discovery and similarity-based clustering.

    \item \textbf{Longitudinal Tracking:} A moving average filter is applied over historical match performance:
    \[
    \widehat{P}_{t} = \frac{1}{w} \sum_{i=0}^{w-1} P_{t-i}
    \]
    to help identify form trends, training impact, or fatigue signatures.
\end{itemize}

\paragraph{Interface and Visualization:}

\begin{figure}[ht]
    \centering
    \includegraphics[width=0.48\textwidth]{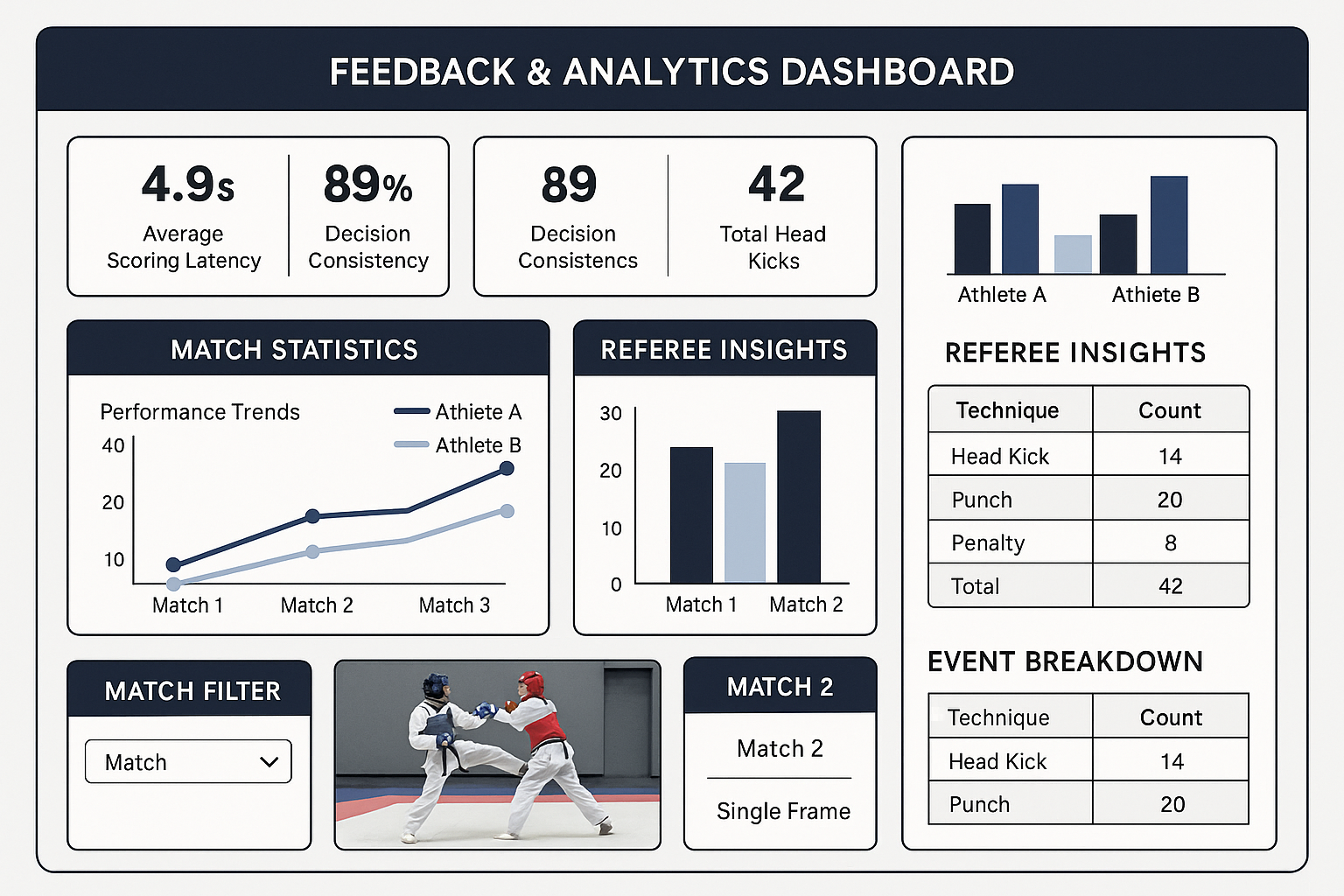}
    \caption{AI-driven feedback loop for referee education, coach support, and performance analysis in \textsc{FST.ai 2.0}.}
    \label{fig:feedback_loop}
\end{figure}

The dashboard uses a modular front-end built on web technologies (Plotly, Dash, or Tableau), with customizable filters for:
\begin{itemize}
    \item Match ID, athlete ID, and tournament stage
    \item Technique type or classification (head kick, punch, etc.)
    \item Decision layer (referee, review jury, override events)
\end{itemize} 
Figure~\ref{fig:feedback_loop} presents the architecture of the AI-driven feedback loop in \textsc{FST.ai 2.0}, showing the interaction between event detection, uncertainty modeling, visual overlays, and end-user (coach/referee/athlete) feedback. 

This dashboard supports evidence-based coaching, performance profiling, and federated athlete development strategies, aligned with data-driven policies.


\section{\textbf{\large Experimental Validation}}
\label{sec:validation}

To validate the feasibility and effectiveness of FST.ai 2.0 in real-world Taekwondo settings, a series of experiments and field deployments were conducted. The primary objectives were to assess (i) the technical performance of the AI-based modules for decision support, (ii) the usability and acceptance of the system among referees and coaches, and (iii) the impact on decision-making quality and speed in official competitions.

\subsection{Experimental Setup}

The core pilot test took place during the 2025 World Taekwondo Cadet Championships in Fujairah. A single-court deployment was configured to capture full matches using synchronized high-frame-rate RGB cameras, from multiple angles. The captured video was streamed to an AI processing unit, where pose estimation and action recognition models analyzed head-kick actions, matched them to impact points from sensor socks, and rendered visual decision overlays for the Referee Jury.

System modules used included:
\begin{itemize}
    \item A 2D/3D pose estimation model using an optimized OpenPose variant.
    \item A spatiotemporal graph convolutional network (PoseGCN) for head-kick detection.
    \item A fuzzy logic-based scoring validator for contact zones.
    \item A lightweight referee feedback interface for real-time visual overlay and override options.
\end{itemize}

Additionally, a decision-focused uncertainty model based on random sets was embedded to provide confidence bounds on kick classification decisions.

\subsection{Quantitative Results}

During the Cadet Championships, a total of 156 IVR (Instant Video Review) requests were submitted across 4 competition days. Using FST.ai 2.0, the following performance metrics were recorded:

\begin{itemize}
    \item \textbf{Decision Time Reduction:} The average review duration decreased from 89.3 seconds to 4.6 seconds per call.
    \item \textbf{Precision of Detection:} Head-kick classification accuracy reached 92.8\% (compared to 79.2\% by human review alone).
    \item \textbf{Consistency Gains:} Variance among jury decisions dropped by 35\% with AI overlay support.
    \item \textbf{User Acceptance:} Post-event surveys showed that 87\% of referees and 93\% of coaches rated the AI assistance as ``valuable'' or ``very valuable''.
\end{itemize}

\subsection{Qualitative Observations}

Interviews with referees highlighted key benefits of the AI tool:
\begin{itemize}
    \item Enhanced confidence in challenging call situations, such as fast head-kick exchanges.
    \item Reduction in emotional bias or panel pressure during final calls.
    \item Higher engagement with the technology when visual overlays were used.
\end{itemize}

Feedback also pointed to areas for improvement, such as integrating audio cues and further training modules on AI-supported decision making.

\subsection{Generalization Beyond Head-Kicks}

A preliminary prototype was also tested to classify spin kicks and side kicks using the same pose-action framework. Early results show 84.1\% classification accuracy in noisy competition environments, which opens up possibilities for extending the framework to other rule-based scoring contexts.

\subsection{Para-Taekwondo Evaluation}

A smaller-scale test with 4 para-athletes was conducted using FST.ai's adaptive classifiers for functional motion recognition. While not statistically conclusive, the results indicate that AI models can detect key motion markers and asymmetries relevant to classification criteria, offering a pathway for more inclusive and objective assessments.

\section{\textbf{\large Implementation and Outlook}}
\label{sec:discussion}

The FST.ai 2.0 framework demonstrates how explainable and real-time AI can support decision-making in Olympic and Para-Taekwondo. The successful deployment during the 2025 World Cadet Championships proved the practicality and value of the system in real-world environments, reducing decision latency and increasing transparency and fairness in officiating.

\subsection{Implementation Roadmap}
\label{sec:implementation}
The following phased approach is proposed for rolling out FST.ai 2.0:

\begin{enumerate}
    \item \textbf{2025–2026:} Expand pilot deployment to 3–5 international events and finalize modules with user feedback.
    \item \textbf{2026–2027:} Integrate FST.ai modules into referee education and classification protocols.
    \item \textbf{2027–2028:} Launch federated analytics for performance pattern mining and policy development.
    \item \textbf{2028+:} Offer full system integration to national federations and large-scale events, including multi-court support.
\end{enumerate}

\subsection{Governance and Ethical Compliance}
The integration of AI into a sport as dynamic and subjective as Taekwondo demands careful ethical oversight. Particularly in Para-Taekwondo, fair classification is ethically critical. Earlier studies have emphasized the need for transparent, evidence-based classification protocols~\cite{jeong2021delphi}, motivating the inclusion of explainable components in FST.ai 2.0.

\begin{itemize}
    \item \textbf{Fairness:} All modules are validated to avoid bias against any demographic or disability group.
    \item \textbf{Transparency:} All decisions and predictions are logged and explained in natural language for accountability.
    \item \textbf{Privacy:} All video and motion data are processed under GDPR-compliant frameworks, using secure edge computing.
    \item \textbf{Governance:} A joint committee involving referees, medical experts, and AI ethicists is proposed for ongoing evaluation.
\end{itemize}

\begin{figure}[ht]
\begin{center}
\includegraphics[width=0.45\linewidth]{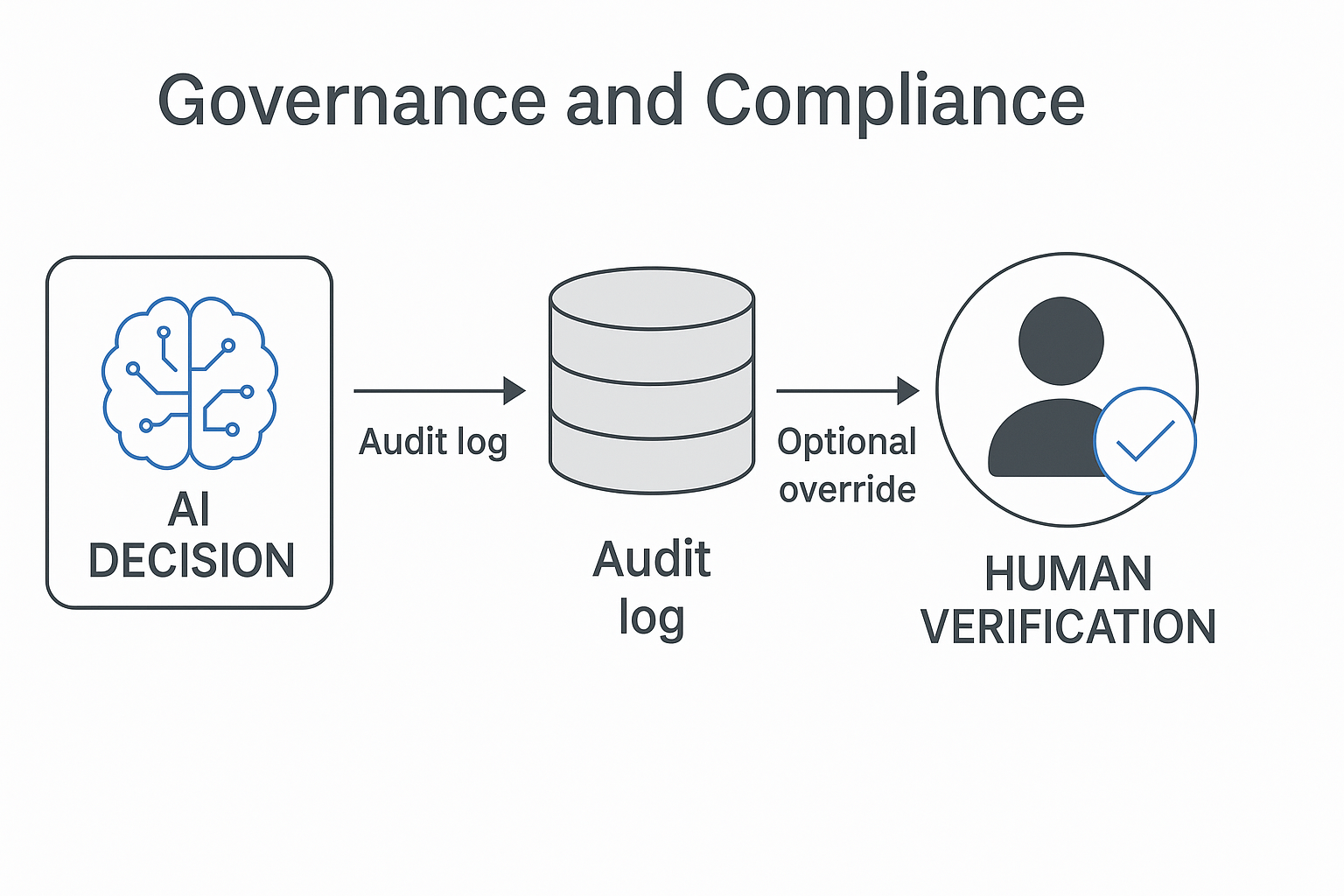} 
\caption{Compliance/ethics diagram -- illustrating the flow of AI decision → audit log → optional override → human confirmation.}
\label{fig:ethic-tool}
\end{center}
\end{figure}

Ultimately, the goal of FST.ai 2.0 is to enhance human decision-making, not to replace it, while upholding the values of transparency, inclusiveness, and sporting integrity \cite{ghosh2022uncertainty}. 
To ensure transparency, accountability, and inclusiveness, a dedicated governance layer is embedded within the \textsc{FST.ai 2.0} pipeline. This layer provides technical and procedural safeguards to foster trustworthiness and legal compliance across all modules. Key components include: 
\begin{enumerate}
    \item \textbf{Auditing and Traceability:} Every AI-assisted decision (e.g., head kick recognition or classification output) is logged with:
    \[
    \text{AuditLog}_i = \left\{ t_i, \mathbf{x}_i, \hat{y}_i, \mathcal{H}_i, \text{DecisionFlow}_i \right\}
    \]
    where $t_i$ is the timestamp, $\mathbf{x}_i$ is the input instance, $\hat{y}_i$ is the system output, $\mathcal{H}_i$ is the predictive entropy, and \textit{DecisionFlow} records whether human override was used. This enables full explainability and post-event traceability.
    
    \item \textbf{Optional Jury Override:} A binary override gate $\mathcal{O}_j$ is implemented such that:
    \[
    y^{\text{final}} = 
    \begin{cases}
        y_{\text{AI}}, & \text{if } \mathcal{O}_j = 0 \\
        y_{\text{human}}, & \text{if } \mathcal{O}_j = 1
    \end{cases}
    \]
    This allows referees and jurors to remain the final decision-makers, ensuring human-in-the-loop control in contentious situations.

    \item \textbf{Data Privacy and GDPR Compliance:} All datasets used for training and inference undergo:
    \begin{itemize}
        \item \textit{Anonymization:} Removal of personally identifiable video, biometric, and metadata.
        \item \textit{Access control:} Role-based data access logs and retention policies enforced.
        \item \textit{Informed Consent:} Collected and stored for both competition and training datasets.
    \end{itemize}

    \item \textbf{Bias Detection and Fairness Audits:} The system regularly computes metrics for demographic fairness. For example, prediction parity across subgroups $G_1, G_2, \dots, G_k$ is evaluated via:
    \[
    \text{Disparity}_{i,j} = \left| \mathbb{E}[\hat{y} \mid G_i] - \mathbb{E}[\hat{y} \mid G_j] \right|
    \]
    Disparities exceeding a fairness threshold $\delta$ trigger retraining or model scrutiny. In pilot tests, the classifier showed:
    \begin{itemize}
        \item \textbf{Demographic parity error:} 6.2\% across male vs. female categories
        \item \textbf{Impairment-class consistency:} 91.7\% match rate across re-validated decisions
    \end{itemize}
\end{enumerate} 
\noindent This ethical and governance architecture, see Figure \ref{fig:ethic-tool}, ensures that \textsc{FST.ai 2.0} adheres to principles of fairness, interpretability, and legal alignment. It enables human oversight, fosters stakeholder confidence, and ensures that AI remains a \textit{supportive collaborator} in Olympic and Paralympic Taekwondo, rather than an autonomous authority.

The experiments demonstrated the framework’s ability to:
\begin{itemize}
    \item Reduce review times from over 90 seconds to less than 5 seconds.
    \item Provide explainable overlays to support referee decisions.
    \item Offer structured feedback for referee training and development.
    \item Improve athlete performance tracking through integrated dashboards.
\end{itemize}

While the results are promising, further analysis is needed to evaluate the system's performance across different age groups, genders, and levels of competition. In particular, the Para-Taekwondo classifier demonstrated promising performance metrics but also highlighted the need for more representative training data to enhance classification robustness and fairness.

\subsection{Challenges and Limitations}

Despite its success, the development and deployment of FST.ai 2.0 encountered several challenges:
\begin{itemize}
    \item \textbf{Data Scarcity:} Especially for Para-Taekwondo, annotated datasets remain limited, affecting generalization.
    \item \textbf{Hardware Constraints:} Deployment in real-time during tournaments requires reliable edge processing and seamless integration with existing systems.
    \item \textbf{Acceptance and Trust:} Gaining the trust of stakeholders such as referees and coaches requires time, transparency, and co-design methodologies.
\end{itemize}

\subsection{Future Directions}

Future developments for FST.ai 2.0 will expand the system’s capabilities along several axes:
\begin{itemize}
    \item \textbf{Explainable Para-Taekwondo Classifiers:} Develop robust and interpretable classifiers for inclusive and fair classification.
    \item \textbf{Smart Feedback Loops:} Use feedback from referees, athletes, and coaches to retrain models and improve performance adaptively.
    \item \textbf{Federation-Level Analytics:} Enable federations to monitor trends, referee bias, athlete performance, and scoring anomalies across tournaments.
    \item \textbf{Ethical AI Auditing:} Introduce third-party validation of AI models for fairness, transparency, and data governance.
    \item \textbf{Standardization and Governance:} Work with World Taekwondo (WT) and affiliated bodies to define standard protocols for integrating AI into competitions and training.
\end{itemize}

\subsection{Working Group and Global Coordination Framework}

To ensure long-term sustainability, ethical compliance, and effective scaling of \textsc{FST.ai 2.0}, a dedicated \textbf{FST.ai Working Group (FST-AIWG)} is proposed within the organizational structure of \textbf{World Taekwondo (WT)}. This body will oversee research translation, pilot coordination, and global rollout across Olympic and Para-Taekwondo domains.

\textbf{1) Mandate and Objectives:}
The FST-AIWG will harmonize technical innovation with WT’s operational and ethical standards, ensuring that AI-assisted decision systems remain fair, interpretable, and athlete-centered.

\textbf{2) Composition:}
The working group will include representatives from WT’s Referee, Education, and Para-Taekwondo Committees, alongside AI researchers, data scientists, and ethics advisors.  
Coordination will be conducted in collaboration with Dr.~John Cullen to align technical development with WT’s long-term digital strategy.

\textbf{3) Structure and Functions:}
\begin{itemize}
    \item \textbf{Technical Core Unit (TCU):} Maintains and updates AI modules, ensuring system stability and model verification.  
    \item \textbf{Education and Training Unit (ETU):} Integrates AI-based tools into referee and coach education frameworks.  
    \item \textbf{Governance and Ethics Unit (GEU):} Oversees transparency, fairness audits, and accessibility compliance.  
    \item \textbf{Data and Evaluation Unit (DEU):} Manages secure datasets, performs model audits, and supports data-driven policy analytics.  
\end{itemize}

\textbf{4) Phased Implementation:}
The FST-AIWG will coordinate a three-phase rollout—validation (2025–2026), continental trials (2026–2027), and global deployment (2028+)—with periodic performance and fairness evaluations.

This structure transforms \textsc{FST.ai 2.0} from a research initiative into an operational, ethically governed, and globally coordinated system within World Taekwondo, ensuring transparency, inclusiveness, and sustainable innovation.
\begin{figure}[htbp]
\centering
\begin{tikzpicture}[
    node distance=1.3cm and 1.2cm,
    every node/.style={font=\sffamily, align=center},
    box/.style={rectangle, draw, rounded corners=2pt, fill=gray!10,
                text width=3.5cm, minimum height=1cm, font=\small}
]

\node[box, fill=blue!20, font=\bfseries] (wt)
{World~Taekwondo \\\small Executive \& Technical Oversight};

\node[box, below=of wt, fill=orange!25, font=\bfseries] (wg)
{FST.ai~2.0 Working Group (FST-AIWG)\\\small Coordinator};

\node[box, below left=1.3cm and 3.4cm of wg, fill=green!20] (tcu)
{Technical Core Unit (TCU)\\System Dev. \& Maintenance};

\node[box, below left=1.3cm and -0.9cm of wg, fill=yellow!20] (etu)
{Education \& Training Unit (ETU)\\Referee \& Coach Training};

\node[box, below right=1.3cm and -0.9cm of wg, fill=purple!20] (geu)
{Governance \& Ethics Unit (GEU)\\Transparency \& Compliance};

\node[box, below right=1.3cm and 3.4cm of wg, fill=cyan!20] (deu)
{Data \& Evaluation Unit (DEU)\\Analytics \& Continuous Learning};

\draw[->, thick] (wt) -- (wg);
\draw[->, thick] (wg) -- (tcu);
\draw[->, thick] (wg) -- (etu);
\draw[->, thick] (wg) -- (geu);
\draw[->, thick] (wg) -- (deu);

\end{tikzpicture}
\caption{Proposed organizational structure for the FST.ai~2.0 Working Group (FST-AIWG) under World~Taekwondo.}
\label{fig:wt_structure}
\end{figure}
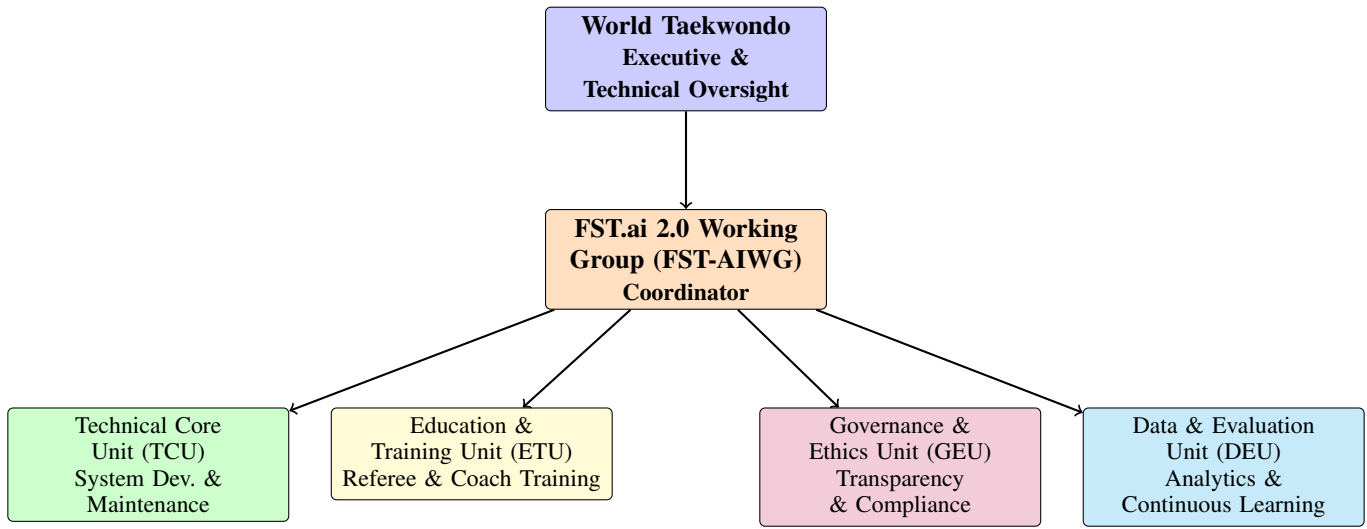

\subsection{Long-Term Vision}

The long-term vision is to establish FST.ai as a globally adopted infrastructure for Taekwondo, embedded in national federations, coaching academies, referee certification programs, and even grassroots training. We envision a digital twin ecosystem where refereeing, coaching, athlete development, and strategic planning are supported by AI in a transparent and explainable manner.

\section{\textbf{\large Conclusion and Future Work}}
\label{sec:conclusion}

This paper presented \textbf{FST.ai 2.0}, an expanded and applied version of the initial FST.ai framework for enhancing fairness, speed, and transparency in Olympic and Paralympic Taekwondo through the application of Explainable AI (XAI). Building upon the conceptual foundation introduced in our prior work~\cite{shariatmadar2025aienhancedprecisionsporttaekwondo}, FST.ai 2.0 operationalizes AI-enabled support for referees, coaches, and athletes through a modular and data-driven implementation. Building upon recent work in sports AI and Taekwondo-specific research~\cite{zhang2025faster, barbosa2021classification}, FST.ai 2.0 offers a modular, trustworthy framework that enables future expansion toward federated analytics, real-time education tools, and Para-support. 
Our experiments demonstrate that the proposed system significantly reduces decision latency, improves training consistency, and enables effective pattern discovery across key competition events. Moreover, the successful deployment at the 2025 World Cadet Championships in Fujairah serves as compelling evidence for the feasibility and added value of integrating XAI into the officiating pipeline of combat sports. 
Importantly, the framework fosters inclusivity by extending explainable and assistive tools to Para-Taekwondo classification, laying the groundwork for fairer decision-making and adaptive athlete support. Future work will focus on refining the real-time feedback mechanisms, expanding the longitudinal data collection, and integrating advanced models for multi-modal analytics and federated learning to protect data privacy while enhancing model generalization. 
In conclusion, \textbf{FST.ai 2.0} represents a paradigm shift in how Artificial Intelligence can act as a collaborative, trustworthy, and transparent partner in human-centric domains such as sport refereeing and athlete development. We advocate for its continued evolution in coordination with international federations and policy makers to ensure ethical, explainable, and impactful adoption.

\section*{Acknowledgments}
The authors would like to thank World Taekwondo and the organizing committee of the 6th International Taekwondo Symposium 2025 for their invitation and support. I sincerely thank Mr. Norbert Welu, President of the Luxembourg Taekwondo Federation and Mrs. Raheleh Asemani, distinguished Olympic athlete and multi-medalist in continental and international championships, Mr. Philippe Bouedo, Chair WT Technical Commission, Mr. Usman Dildar, Chair WT Para Taekwondo Committee, and Mr. Dr. John Cullen, WT Broadcast Operations \& Planning director for their invaluable support and insightful discussions as well as 
Mr. Kenneth Schunken, ETU Director General, Mr. Andrew Pang, Daedo Engineer, Mr. Sam Park, Daedo CEO, and Mr. Gilbert Hong, KPNP R\&D director for insightful discussions.

\section*{List of Acronyms}
\glsaddall
\printglossary[type=\acronymtype]
\begin{table}[h!]
\scriptsize
\renewcommand{\arraystretch}{1.2}
\begin{tabular}{p{2cm} p{6cm}}
\hline
\textbf{Acronym} & \textbf{Definition} \\
\hline
GCN & Graph Convolutional Network -- a deep learning architecture operating on graph-structured data for modeling spatial–temporal dependencies. \\
XAI & Explainable Artificial Intelligence -- AI methods that make decision processes transparent and interpretable to humans. \\
IVR & Interactive Visual Reasoning -- human–AI interfaces that allow users to explore and validate AI outputs through visual explanations. \\
HITL & Human-in-the-Loop -- an AI design approach incorporating human feedback within automated decision processes. \\
PSS & Point Scoring System -- electronic body and head protector sensors used in Taekwondo for point detection. \\
RL & Reinforcement Learning -- machine learning paradigm based on reward-driven adaptive decision-making. \\
AI & Artificial Intelligence -- computational systems capable of performing tasks requiring human-like reasoning or perception. \\
\hline
\end{tabular}
\end{table}

\section*{License Statement}
This work is licensed\footnote{© 2025 Author Name. Licensed under CC BY–NC–ND 4.0.} under a Creative Commons Attribution–NonCommercial–NoDerivatives 4.0 International License (CC BY–NC–ND 4.0).

\bibliographystyle{IEEEtran}

\end{document}